\documentclass[nonacm,acmtog,balance=false]{acmart}

\setcopyright{none}
\copyrightyear{2025}
\acmYear{2025}
\acmDOI{XXXXXXX.XXXXXXX}

\setcitestyle{nosort,square}

\usepackage{microtype} 
\usepackage[nosort,capitalize]{cleveref}
\usepackage{mathtools}
\usepackage{enumitem}
\usepackage{nicefrac}
\usepackage{bm}
\usepackage{booktabs}
\usepackage{bbm}
\usepackage[normalem]{ulem}

\newcommand{\surfrv}{S}
\newcommand{\avgsurf}{\bar S}

\newcommand{\Li}{L_\mathrm{i}}
\newcommand{\Lie}{L_\mathrm{i}^{\hspace{-.3mm}\varepsilon}}

\newcommand{\Lo}{L_\mathrm{o}}
\newcommand{\LoS}{L_\mathrm{o}^\mathrm{S}}
\newcommand{\Lobg}{L_\mathrm{o}^\mathrm{\bar S}}

\newcommand{\Libg}{L_\mathrm{i}^\mathrm{\bar S}}
\newcommand{\Ls}{L_\mathrm{s}}
\newcommand{\Le}{L_\mathrm{e}}
\newcommand{\mut}{\mu_\mathrm{t}}
\newcommand{\mus}{\mu_\mathrm{s}}
\newcommand{\mua}{\mu_\mathrm{a}}

\newcommand{\vx}{\mathbf{x}}
\newcommand{\vw}{\bm{\omega}}
\newcommand{\vy}{\mathbf{y}}

\newcommand{\dd}{\,\mathrm{d}}
\newcommand{\dt}{\dd t}

\newcommand{\dw}{\dd \vw}
\newcommand{\loss}{\mathcal{L}}
\newcommand{\render}{\mathcal{R}}
\newcommand{\params}{\pi}
\newcommand{\Params}{\Pi}
\newcommand{\alt}{\tilde}
\newcommand{\altrender}{\mathcal{\alt R}}
\newcommand{\surfaces}{\mathcal{M}}

\newcommand{\occupancy}{\alpha}
\newcommand{\orientation}{\bm{\beta}}

\newcommand{\implicit}{\Phi}

\newcommand{\x}{\mathbf{x}}
\newcommand{\p}{\mathbf{p}}
\newcommand{\vth}{\mathbf{v}_{\theta}}
\newcommand{\xa}{\mathbf{x}_\mathrm{a}}
\newcommand{\xb}{\mathbf{x}_\mathrm{b}}
\newcommand{\xc}{\mathbf{x}_\mathrm{c}}
\newcommand{\Na}{\mathbf{N}_\mathrm{a}}
\newcommand{\Nb}{\mathbf{N}_\mathrm{b}}
\newcommand{\Nc}{\mathbf{N}_\mathrm{c}}

\newcommand{\tb}{\mathbf{t}_\mathrm{b}}
\newcommand{\tc}{\mathbf{t}_\mathrm{c}}
\newcommand{\mb}{\mathbf{m}_\mathrm{b}}
\newcommand{\lengthab}{l_\mathrm{ab}}
\newcommand{\lengthac}{l_\mathrm{ac}}
\newcommand{\nb}{\mathbf{n}_\mathrm{b}}
\newcommand{\nc}{\mathbf{n}_\mathrm{c}}

\newcommand{\bomega}{\bm{\omega}}

\newcommand{\dl}{\,\mathrm{d}l}

\newcommand{\dmu}{\,\mathrm{d}\mu}

\newcommand{\dn}{\,\mathrm{d}n}
\newcommand{\dA}{\,\mathrm{d}A}
\newcommand{\dV}{\,\mathrm{d}V}

\newcommand{\xk}{\mathbf{x}_\mathrm{k}}
\newcommand{\xkm}{\mathbf{x}_{\mathrm{k}-1}}
\newcommand{\xkp}{\mathbf{x}_{\mathrm{k}+1}}

\DeclareMathOperator{\erf}{erf}
\DeclareMathOperator*{\argmin}{argmin}
\DeclareMathOperator*{\corr}{corr}

\usepackage{tcolorbox}
\usepackage[frozencache,cachedir=.]{minted}
\usepackage{mdframed}
\usepackage{setspace}
\definecolor{hlcol}{RGB}{255,200,200}
\newminted{cpp}{autogobble,escapeinside=||,fontsize=\footnotesize}
\newminted{python}{autogobble,escapeinside=||,fontsize=\footnotesize}

\BeforeBeginEnvironment{minted}{
   \begin{tcolorbox}[arc=1pt,colback=black!3!white,colframe=black!16!white,boxrule=0.3mm,before skip=1mm,after skip=1mm,top=.5mm,bottom=.5mm,left=1mm,right=1mm]
   }
\AfterEndEnvironment{minted}{\end{tcolorbox}}
\AtBeginEnvironment{minted}{\setstretch{1.2}}

\begin{document}
\title{Many-Worlds Inverse Rendering}

\author{Ziyi Zhang}
\affiliation{
   \institution{École Polytechnique Fédérale de Lausanne (EPFL)}
   \city{Lausanne}
   \country{Switzerland}
}

\author{Nicolas Roussel}
\affiliation{
   \institution{École Polytechnique Fédérale de Lausanne (EPFL)}
   \city{Lausanne}
   \country{Switzerland}
}

\author{Wenzel Jakob}
\affiliation{
   \institution{École Polytechnique Fédérale de Lausanne (EPFL)}
   \city{Lausanne}
   \country{Switzerland}
}

\renewcommand{\shortauthors}{Zhang et al.}

\begin{teaserfigure}
    \centering
    \includegraphics[width=\textwidth]{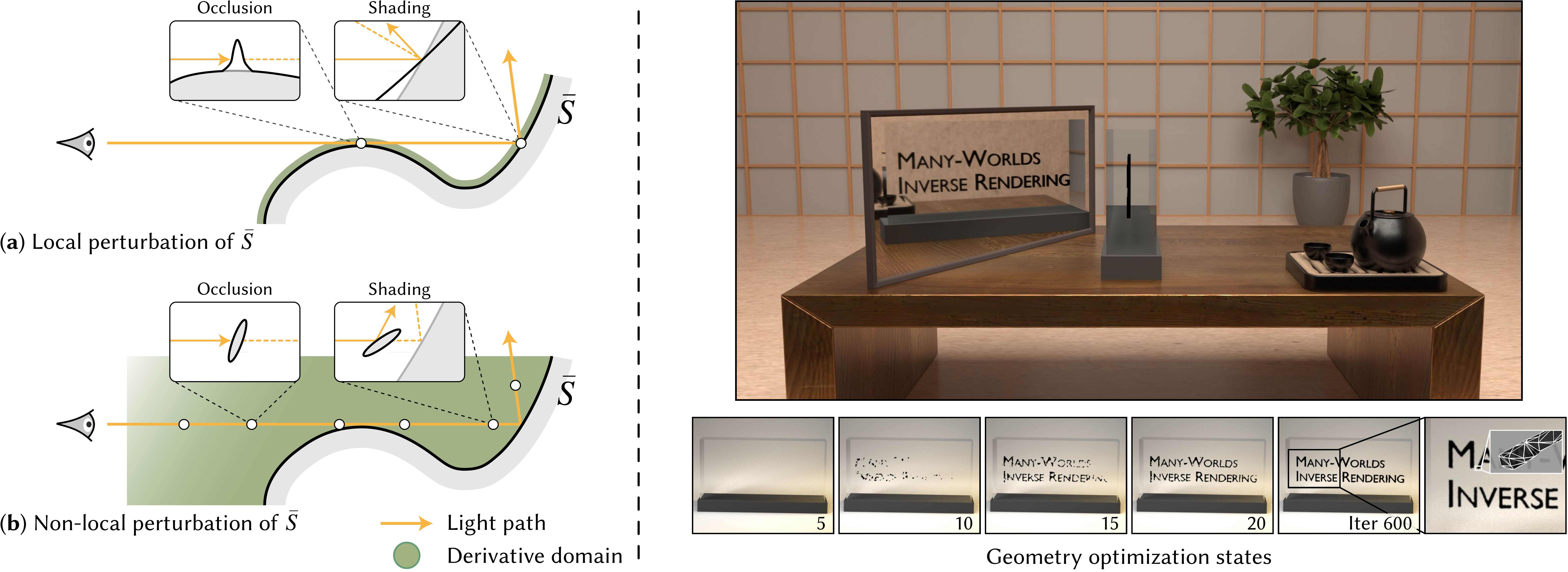}
      \vspace{-6mm}
    \caption{%
        \label{fig:teaser}%
        \textbf{Left:}  Unlike methods that compute local surface derivatives, as shown in (a),
        our method models a more general non-local surface perturbation, as shown in (b).
        By treating \emph{potential} surface patches as \emph{conflicting} perturbations of a background surface $\avgsurf$, we extend derivative propagation into space.
        This allows geometry to emerge in empty regions, with convergence behavior similar to volume rendering---yet without requiring multiple scattering simulations or approximating volumetric appearance using surface materials.
        \textbf{Right:} An example use of \emph{many-worlds derivatives} reconstructing a triangle mesh embedded in a glass block and indirectly observed through a mirror.
    }
\end{teaserfigure}

\begin{abstract}
    Discontinuous visibility changes remain a major bottleneck when optimizing surfaces within a physically based inverse renderer.
    Many previous works have proposed sophisticated algorithms and data structures to sample visibility silhouettes more efficiently.

    Our work presents another solution: instead of differentiating a tentative surface locally, we differentiate a non-local perturbation of a surface.
    We refer to this as a ``many-worlds'' representation because it models a non-interacting superposition of conflicting explanations  (``worlds'') of the input dataset.
    Each world is optically isolated from others, leading to a new transport law that distinguishes our method from prior work based on \mbox{exponential random media.}

    The resulting Monte Carlo algorithm is simpler and more efficient than prior methods.
    We demonstrate that our method promotes rapid convergence, both in terms of the total iteration count and the cost per iteration.
\end{abstract}

\maketitle


\section{Introduction}
\label{sec:intro}
Dramatic progress in the area of inverse rendering has led to methods that can fully reverse the process of image formation to reconstruct 3D scenes from 2D images.

This is usually formulated as an optimization problem: given a loss function $\loss$ and a rendering $\render(\params)$ of tentative parameters $\params$, we seek to minimize their composition
\begin{equation}
    \label{eqn:inverse-rendering}
    \params^{*}=\argmin_{\params\in\Params}\loss(\render(\params)).
\end{equation}
The details greatly vary depending on the application, but usually $\loss$ will quantify the difference between the rendering and one or more reference images.

Recent work on this problem is largely based on \emph{emissive} volume reconstruction~\cite{NeRF,GaussianSplats}.
The term ``emissive'' highlights that this approach does not rely on simulating light physics or material reflectance;
instead, the volume is treated as if it was a natural emitter of light and stores color values representing this emitted radiation.
These methods are popular because the direct mapping between stored color and observed appearance results in well-behaved optimization problems.

Physically based methods instead seek a more complicated explanation: they simulate emission and scattering inside a general scene that can contain essentially anything: surfaces, volumes, physically based BRDFs, etc.
Interreflection adds dense nonlinear parameter dependencies that make this a significantly harder optimization problem.
It goes without saying that these two approaches do not compete: when an emissive volume is an acceptable answer, it should always be preferred.
This is because it encapsulates material, lighting, geometry and inter-reflection in a unified field representation \mbox{that promotes speedy and stable convergence.}

That said, many applications rely on reconstructions that account for indirect cues---such as shadows and interreflections---and aim to produce physically meaningful results suitable for downstream tasks like relighting and editing.
Unfortunately, algorithms designed for this harder physical problem are often \emph{surprisingly brittle}.

A default strategy is to use gradient-based descent to evolve a scene representation (e.g., a triangle mesh) in a domain $\Params$ containing the target $\params^*$.
It seems only natural that we try to reach this goal by evolving compatible models \mbox{$\pi_i\in\Params$} in this domain, e.g., by deforming a tentative surface starting from an initial guess.




However, in physical light simulation,
computing $\nabla_\params\loss(\render(\params_i))$ with automatic differentiation produces incorrect gradients due to parameter-dependent discontinuities. 
Incorporating specialized techniques to fix this problem simply leads to the next one: the algorithm now outputs sparse gradients on object silhouettes that cause convergence to bad local minima.
Adding further precondition and regularization can help.
But even with all of these fixes in place, the optimization often still does not work all that well.


Instead, we consider optimizations on an extended parameter space:
\begin{equation}
    \vspace{-.5mm}
    \label{eqn:alternative-optimization}
    \params^*, \alt\params^*=\argmin_{\params\in\Params,\,\alt\params\in\alt\Params}\mathcal{L}(\altrender(\params, \alt\params)),
\end{equation}
where $\alt\render$ is furthermore parameterized by $\alt\params\in\alt\Params$ representing features that we are \emph{unwilling to accept} in the final solution.
If the role of these ``perpendicular dimensions'' diminishes over time,
then we can simply discard $\alt\params^*$ at the end and take $\params^*$ to be the solution of the original optimization problem.

The parameter space extension serves two purposes: first, it turns a circuitous trajectory through a non-convex energy landscape into a more direct route by using the extra degrees of freedom:
\begin{center}
    \vspace{-0mm}
    \includegraphics[width=.7\columnwidth]{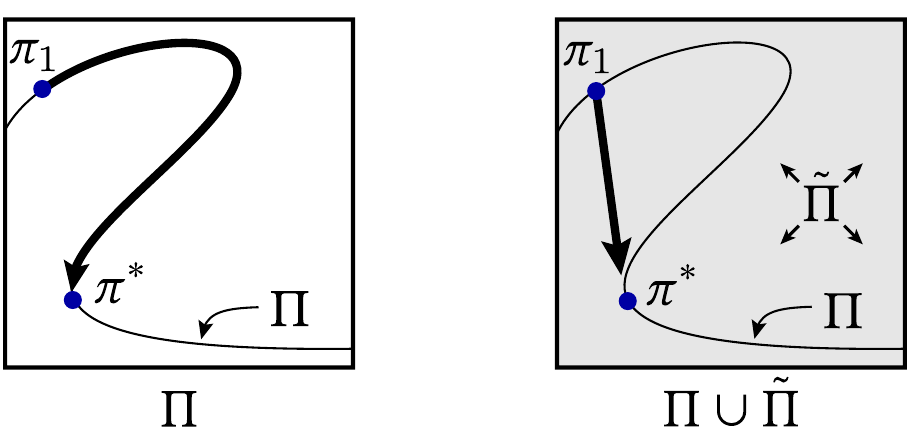}
    \vspace{-1mm}
\end{center}%
Second, we will choose the extension so that visibility changes in the original problem cease to be discontinuous in the extended space, which greatly reduces algorithmic complexity.

Seen from a high level, this isn't a new idea:
numerous works in emissive surface reconstruction \cite{Yariv2021Volume,wang2021neus,Miller2023Theory}
have shown that the problem becomes tractable when starting from an emissive volume like NeRF.
However, taking this idea to the world of physically based rendering (e.g., by optimizing a scattering microflake volume) leads to several serious problems:

\begin{enumerate}[leftmargin=6.3mm]
    \item \textbf{Speed}.
    Introducing random volumes into a physically based renderer requires solving the \emph{radiative transfer equation}~(RTE), which involves free-flight sampling and transmittance estimation along every ray segment.
    In heterogeneous volumes, both are computationally expensive iterative processes, making this approach significantly slower than surface-based methods.

    \item \textbf{Extraction}.
    A separate optimization is required to convert converged volumes into optically similar surfaces with BRDFs, and the resulting surface is often only an approximation of the volume appearance.
    Prior work related the bulk properties of microfacets and microflakes~\cite{dupuy2016additional,Miller2023Theory}, but this insight cannot be used to convert a general microflake volume into a surface, nor does it generalize to the richness of reflectance models in modern rendering systems.
\end{enumerate}

Instead of introducing an exponential volume that blends all possible surfaces multiplicatively,
we retain a surface (denoted $\avgsurf$) and augment it with a spatial representation $S$ to model \emph{non-local perturbations} of that surface.

Consider placing a new surface patch above the original surface.
Standard differentiation of a path tracer does not account for this type of change, yet it is a valid way to evolve the scene representation---one that evolves the surface in a non-local manner.

Importantly, these non-local perturbations can occur simultaneously along a light path:
the optimization of a potential surface patch at one location is independent of that at another location.
This is because these hypothetical surfaces \emph{aren't real} (or at least, not yet).
They constitute different surface possibilities that may or may not become part of the final reconstruction.
It makes little sense for them to affect each other.

This leads to a new transport law that we refer to as \emph{many-worlds derivative transport}.
The term ``many-worlds'' not only suggests a distribution of surfaces but also emphasizes the \emph{non-interacting} nature of different perturbations (``worlds'').

Differentiation of this model produces dense surface derivatives in an extended domain.
Since we avoid diffusing values into an exponential volume, the optimization remains as efficient as surface evolution methods.
Finally, mesh extraction is a simple projection that discards $S$, and all of this is easily incorporated into a physically based path tracer.

Although this paper sometimes refers to the spatial representation $S$ as a ``volume'', our method should \emph{not} be interpreted as volume reconstruction.
Specifically, we never optimize this ``volume'' $S$ to match input images; instead, we use it to refine the surface $\avgsurf$.
Moreover, our method interacts with BRDFs rather than phase functions, and has no concept of transmittance in its radiative transfer.

Our main contribution is a solution to the classic discontinuity problem that is algorithmically simpler and computationally more efficient than prior methods.
We present two derivations that alternatively formulate the central algorithm as either
\begin{itemize}[leftmargin=6.3mm]
    \item a non-local perturbation of an evolving surface (\autoref{sec:method}), or
    \item an extension of the surface derivative domain that removes the need for silhouette sampling (\autoref{sec:discussion} and \autoref{sec:appendix_derivation}).
\end{itemize}


\section{Related work and background}
\label{sec:related}
This section reviews related prior work on inverse rendering of solid geometry.
The simultaneous disentanglement of geometry, material and lighting is orthogonal to the topic of this work.

\subsection{Physically based inverse rendering}
The physically based approach tries to model all available information in input images by accounting for lighting, materials, and interreflection.
This leads to several challenges.

\paragraph*{Discontinuities}
Analytic differentiation of the rendering equation produces a boundary derivative term that is not sampled by primal rendering strategies.
Consequently, applying automatic differentiation to a rendering algorithm produces incorrect gradients.

Estimating the missing derivative requires sampling rays along shape boundaries (i.e., silhouettes).
Prior works \cite{li2018differentiable,Zhang2020PSDR,yan2022efficient,Zhang2023Projective,Xu2023PSDR-WAS,Xu2024PSDR-LMC} have extensively studied the theory and practice of estimating this boundary term.
Another approach employs reparameterizations to convert the boundary contribution to neighboring regions, enabling simpler area-based formulations.
These methods \cite{Loubet2019Reparameterizing,bangaru2020unbiased} often require tracing auxiliary rays to detect the presence of neighboring silhouettes.
For geometry representations based on \emph{signed distance fields} (SDFs), specialized algorithms can alleviate some of this cost \cite{vicini2022differentiable, Bangaru2022NeuralSDFReparam, Wang2024Simple}.

While the theory of the boundary term continues to evolve, practical efficiency and robustness remain challenging.
Current methods still have limitations concerning bias (e.g., specular surfaces) and variance (e.g., when multiple silhouettes are in close proximity).
These methods are difficult to implement, and the cost of boundary handling is usually the main bottleneck of the entire optimization.

\paragraph*{Sparse gradients}
The derivative of the rendering process can be decomposed into two types: the continuous part and the discontinuous part.
We refer readers to Figure~4 in \citeauthor{Zhang2023Projective}'s work \shortcite{Zhang2023Projective} for an illustration.
Since our paper focuses on geometry optimization, we refer to the continuous part as the \emph{shading derivative} and the discontinuous part as the \emph{boundary derivative} in the following.

The shading derivative is defined everywhere on the surface and primarily affects the surface through its normal and heavily depends on the lighting.
Adjoint rendering methods~\cite{NimierDavid2020Radiative,Vicini2021PathReplay} exploit symmetries to compute the shading derivative with linear time complexity.
\citet{Zeltner2021MonteCarlo} analyze different strategies for variance reduction in this process.
The boundary derivative, on the other hand, is only defined on the visibility silhouette of the surface.
These surface derivatives are sparse in two senses:
\begin{enumerate}[leftmargin=6.3mm]
    \item A single view only observes a limited subset of all possible silhouettes, and a derivative step is likely to introduce a kink at this subset.
     \citet{Nicolet2021Large} reparameterize meshes on a space that promotes smoothness to mitigate this issue.
    \item The derivatives are only defined on the surface, not throughout the entire 3D space.
     Although light transport is simulated in this larger space, only the neighborhood of a surface can use the computed gradients.
     Regions far from the initial guess, must wait until the surface extends to them to receive gradients.
     \citet{mehta2023theory} propose an elegant solution for vector graphics in 2D; however, they do not extend their approach to define surface derivatives in the entire 3D space.
\end{enumerate}

The sparsity has two implications: (1) the optimization is slow because it needs many iterations to deform the surface to the target shape, and (2) a good initial guess is needed, especially when the target shape has a complicated topology or self-occlusions.

\paragraph*{Approximations}
There is a substantial body of literature on approximating the PBR process to make optimization easier.
These methods often bypass the discontinuity problem by assuming knowledge of a shape mask (implicitly assuming the shape is directly observable), approximating boundary derivatives \cite{Laine2020diffrast}, or diffusing the rendering process into a higher-dimensional space \cite{fischer2023plateau}.
Additionally, many methods approximate the light transport for efficiency \cite{physg2021,jin2023tensoir,zhang2021nerfactor,boss2021nerd}.
These methods are physically inspired and can handle complex scenes, while our theory focuses on differentiating a fully physically based rendering process.

\citet{NimierDavid2022Unbiased} demonstrate that a physically based non-emissive volume can be optimized to fit the input images.
These methods are more physically accurate than the emissive volume methods but do not yield a surface representation in the end.

\subsection{Emissive volume inverse rendering}
A large body of work based on variants of the NeRF~\cite{NeRF} technique replaces the expensive rendering process with an emissive volume, which leads to
a fog-like volume that does not accurately represent a solid surface.

Later work \cite{Yariv2021Volume,wang2021neus,li2023neuralangelo,Miller2023Theory} added a surface prior to the volume model by introducing an SDF and deriving the volume density from it.
The SDF values are mapped to volume densities using a distribution function, assuming that a ray intersects a stochastic object following a Markov process.
The variance of the distribution is learnable during optimization: it begins with a high-variance distribution, mimicking the NeRF model, and gradually reduces to ensure that a surface can confidently be extracted.

These SDF-parameterized volume reconstructions can be seen as a \textit{blurry} version of surface-based methods without indirect effects.
When the variance is zero, these methods simplify to surface-based methods\footnote{This equivalence is theoretical.
In practice, algorithms designed for volume optimization cannot handle such a Dirac-delta density distribution due to discontinuities.}.
With moderate variance, the effect of a solid object, which we can interpret as ``terminating a ray at a position with probability $1$'', is diffused to the neighborhood, interpretable as ``terminating a ray in this neighborhood according to a probability density''.

\subsection{Surfaces versus volumes}
Several works studied the relationship of surface and volume rendering: \citeauthor{heitz2016multiple}~\shortcite{heitz2016multiple} and \citeauthor{dupuy2016additional}~\shortcite{dupuy2016additional} describe microfacet surfaces within the framework of random volumes.
\citeauthor{vicini2021non}~\shortcite{vicini2021non} incorporated non-exponential transmittance to account for correlations arising from opaque surfaces.
\citeauthor{Miller2023Theory}~\shortcite{Miller2023Theory} formalize microflake volumes as the relaxation of a stochastic surface model and solve inverse problems using this representation.
\citeauthor{seyb2024microfacets}~\shortcite{seyb2024microfacets} recently incorporated a richer set of spatial correlations, producing a forward rendering framework that spans the full continuum ranging from pure volume to surface-like interactions.


\section{Method}
\label{sec:method}

\subsection{Motivation}
\label{Sec:motivation}

\paragraph{Extended parameter space}
Our algorithm evolves a surface $\avgsurf$ derived from a distribution $S$ over potential surfaces.
Rather than modifying the surface $\avgsurf$ directly, we adjust $S$, which defines a density field from which $\avgsurf$ is extracted---typically as a level set.
As the density field changes, so does the geometry of $\avgsurf$, allowing indirect but effective surface optimization.
The details of how these spaces are defined are orthogonal to our method, and we describe them in \autoref{sec:implementation}.

\autoref{fig:teaser}b demonstrates the effect of adding a \emph{hypothetical surface patch} (drawn from $S$) into a scene containing the surface $\avgsurf$.
This patch modifies how light propagates in the scene, thereby changing the surface rendering of $\avgsurf$.
By adjusting the existence and properties (e.g., normals, BRDFs) of such patches, we iteratively improve $\avgsurf$ to better match the target image.

We term this a \emph{non-local perturbation} of $\avgsurf$:
optimizing such hypothetical patches propagates derivatives across the entire domain of $S$, rather than confining updates locally on the surface.

This isn't an entirely different approach from prior work:
in the limiting case where the perturbation position coincides with $\avgsurf$ itself, our derivatives exactly match standard surface derivatives in physical light simulation
[\citeauthor{Zhang2020PSDR}~\citeyear{Zhang2020PSDR}, \citeauthor{Zhang2023Projective}~\citeyear{Zhang2023Projective}] (see \autoref{sec:appendix_derivation} for details on this equivalence).
This shows that our method is a generalization of local surface evolution, extending its domain while preserving its geometric meaning.

Optimization converges when no further perturbation improves the match between $\avgsurf$ and the target image.
At this point, we discard $S$ and keep $\avgsurf$.

Throughout the optimization, $\avgsurf$ serves as a static background, providing base colors for perturbations without being optimized itself.
For this reason, we also refer to $\avgsurf$ as the \emph{background surface}.

\paragraph{Conflicting possibilities}
Consider a simple case of two non-local perturbations along a ray.
We shall think of them as competing candidates for improving the agreement between the radiance from $\avgsurf$ and the reference image.
If the optimization seeks to increase the radiance value along the ray (i.e., if $\frac{\partial \mathrm{loss}}{\partial \Li}$ is positive), the same information should be propagated to both positions without weighting.

\begin{center}
    \includegraphics[width=0.75\columnwidth]{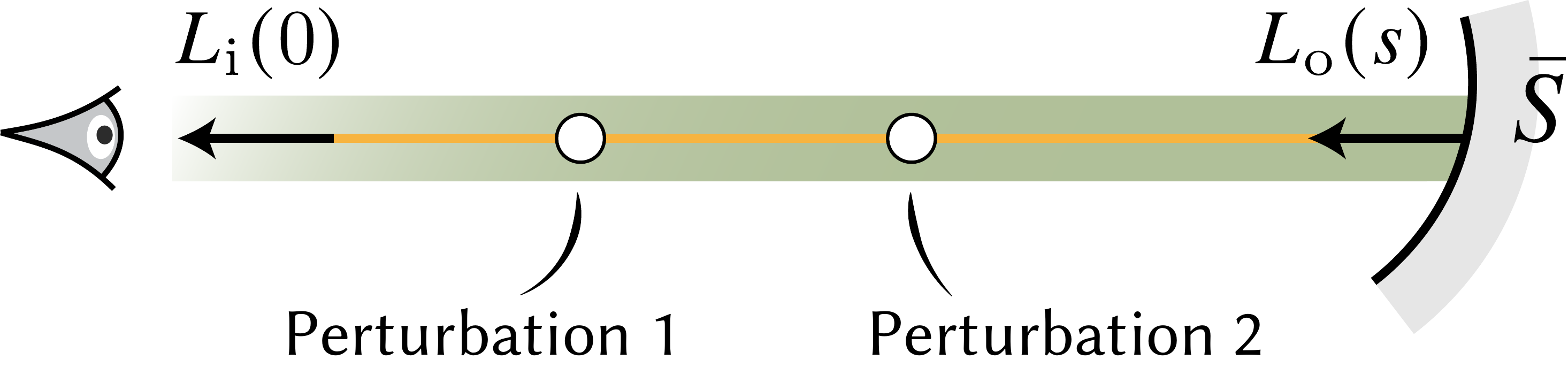}
\end{center}%
Here, it might seem natural to distribute the target update between the two positions --- say, by scaling it by $\frac{1}{2}$.
However, this weighting implicitly assumes the two perturbations \emph{compound} to refine the same surface $\avgsurf$.
In reality, they are \emph{mutually exclusive} possibilities --- once optimization converges, only one will contribute to the final  radiance, without any kind of blending.

This principle extends to more than two perturbations:
all candidate positions along the ray should receive the same target update, as if existing in \mbox{\emph{many worlds}} that do not interact.
Ultimately, the ray will intersect only one of these possibilities, which becomes part of the final surface.

The above discussion explains the motivation behind our method at an intuitive level.
Our next goals are therefore to quantify non-local perturbations and derive the derivative transport law.

\subsection{Many-worlds derivative transport}

\paragraph{Single-world case}
Consider a scene containing only the background surface $\avgsurf$.
Radiance along ray $(\vx, \vy)$ propagating in direction $\vw$ remains constant:
\begin{equation*}
    \Libg(0) = \Lobg(s),
\end{equation*}
where \mbox{$\Libg(0) \coloneqq \Libg(\vy, -\vw)$} (incident radiance at $\vy$) and
\mbox{$\Lobg(s) \coloneqq \Lobg(\vx, \vw)$} (outgoing radiance at $\vx$) are parameterized by distance.

For a non-local perturbation candidate at distance $t$, we model its impact on radiative transport as:
\begin{equation}
    \label{eqn:perturb-render}
    \Li(0)
    = \occupancy(t) \underbrace{\LoS(t)}_\text{\footnotesize perturbed}
    + \left[1 - \occupancy(t)\right] \underbrace{\Lobg(s)}_\text{\footnotesize original},
\end{equation}
where $\alpha(t) \in [0,1]$ is the probability of a hypothetical surface patch existing at $t$.
The perturbed radiance
$\LoS(t)$ computes reflected radiance \textit{as if} the patch were inserted at $t$, while the rest of the scene remains $\avgsurf$.

While this formulation is of little use for physically based rendering, its
derivative provides the means to optimize on the extended parameter space:
\begin{align}
    \label{eqn:perturb-deriv}
    \partial_\pi \Li(0)
    &=
    \underbracket{\partial_\pi \occupancy(t) \, [\LoS(t) - \Lobg(s)]}_{\text{(i) \footnotesize Occlusion}}
    + \underbracket{\occupancy(t) \, \partial_\pi \LoS(t)}_{\text{(ii) \footnotesize Shading}}.
\end{align}
To increase transmitted radiance---that is, if $\partial_\pi \Li(0)$ is positive---we can:
\begin{enumerate}[leftmargin=6.3mm]
    \item Raise the occupancy at $\alpha(t)$ if this perturbation is favorable, i.e., $\LoS(t)>\Lobg(s)$, or lower it otherwise.
    \item Increase the reflected radiance $\LoS(t)$, e.g., by altering the normal or BRDF of the hypothetical surface.
\end{enumerate}
Both operations locally update the distribution $S$ at $t$.

\begin{figure}[t]
    \centering
    \includegraphics[width=1.0\columnwidth]{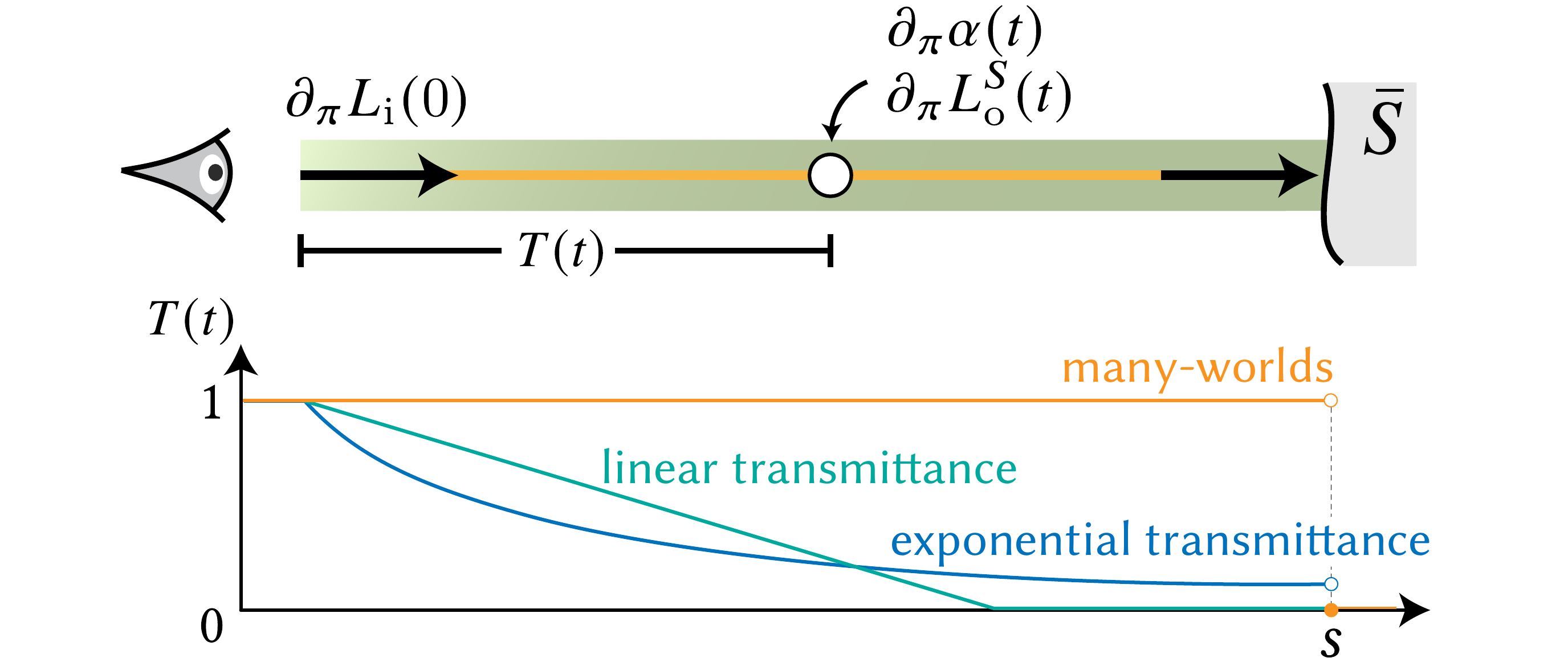}
    \caption{%
        \label{fig:transmittance}%
        \textbf{Many-worlds derivative transport}. The propagated derivative at distance $t$ is only weighted by the local radiance difference and the local occupancy respectively (\autoref{eqn:manyworlds-deriv}).
        In contrast to exponential or non-exponential (e.g., linear~\cite{vicini2021non}) volumes, the notion of \emph{transmittance} disappears as it models nonsensical inter-world shadowing.
        }
\end{figure}

\paragraph{Many-worlds case}
We now extend to the situation where every point along the ray can be considered a potential perturbation.
As discussed in \autoref{Sec:motivation}, we aim to distribute the target update uniformly across all possibilities along one segment.

Similar to basic arithmetic, where a sum
$c=a+b$ propagates the equal derivative to both $a$ and $b$ without weighting ($\frac{\partial c}{\partial a} = 1$, not $\frac{1}{2}$),
we sum radiative contributions from distinct perturbations,
expressed as the integral form of a \emph{radiative transfer equation} (RTE) along the ray:
\begin{equation}
    \label{eqn:manyworlds-render}
    \Li(0)
    =\int_0^s  \Big(  \occupancy(t) \, \!\!\!\!\!\! \underbrace{\LoS(t)}_{\text{candidate at}~t} \!\!\!\!\!\!
    + \,
    \left[1 - \occupancy(t)\right] \, \!\!\! \underbrace{\Lobg(s)}_\text{background} \!\!\! \Big) \dt.
\end{equation}
Differentiating it yields the many-worlds derivative transport law:
\begin{equation}
    \label{eqn:manyworlds-deriv}
    \partial_\pi \Li(0)
    =\int_0^s  \Big(  \partial_\pi \occupancy(t) \, [\LoS(t) - \Lobg(s)] +
     \occupancy(t) \, \partial_\pi \LoS(t)  \Big) \dt.
\end{equation}
The derivative propagation in \autoref{eqn:manyworlds-deriv} lacks transmittance terms that would ordinarily model attenuation along the ray (\autoref{fig:transmittance}).
This arises from the core principle that distinct worlds must not interact.

The only way in which the background surface $\avgsurf$ manifests in this equation is to provide a single baseline radiance value $\Lobg(s)$ needed to compute a difference
of radiance values.
As a result, $\avgsurf$ does not directly receive gradients;
yet, it still evolves during optimization as a result of changes in the distribution $S$ that \mbox{dictates $\avgsurf$}.

\paragraph{Parameterization}
\autoref{eqn:manyworlds-deriv} requires that derivatives be propagated to two quantities in the extended parameter space: the existence probability of a surface patch, and the outgoing radiance $\LoS(\x, \vw)$ determined by the surface patch properties.

\begin{figure}[t]
    \centering
    \includegraphics[width=\columnwidth]{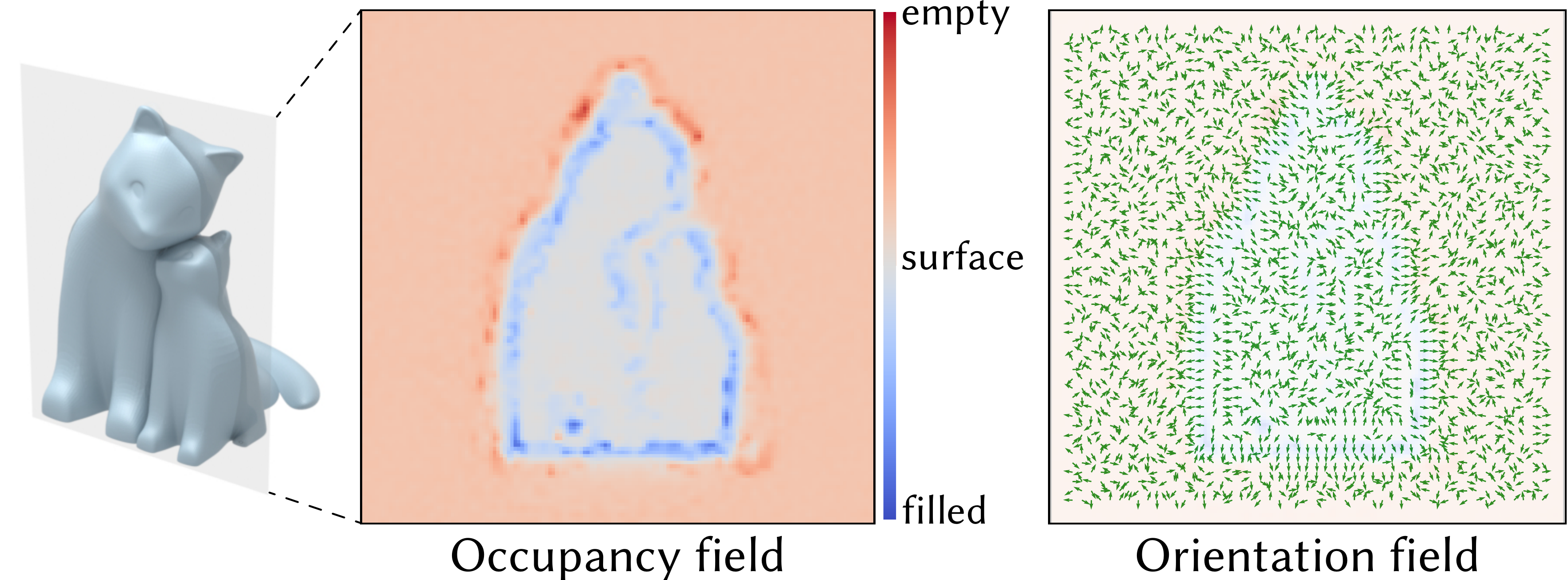}
    \caption{%
        \label{fig:alpha_beta_illustration}%
        \textbf{Occupancy and orientation fields.}
        The above images visualize the contents of an occupancy ($\occupancy$) and orientation ($\orientation$) field following an optimization.
        The former models the probability of a surface existing at a position, while the latter assigns normal directions.}
\end{figure}

Any parameterization of $S$ that supports differentiation of these quantities is suitable.
Our implementation uses an occupancy field \mbox{$\occupancy(\vx):\mathbb{R}^3\to[0,1]$} and an orientation field \mbox{$\orientation(\vx):\mathbb{R}^3\to\mathcal{S}^2$} (\autoref{fig:alpha_beta_illustration}).

The orientation field $\orientation(\vx)$ assigns a normal direction to the surface patch at $\vx$.
The occupancy field $\alpha(\vx)$ \cite{mescheder2019occupancy,niemeyer2020differentiable} models the probability
\begin{equation}
    \label{eqn:occupancy}
    \occupancy(\vx)\coloneqq
    \Pr\{\vx \text{ is inside of } S\},
\end{equation}
and is zero if the patch is interacted from the back side ($\vw \cdot\orientation(\vx)<0$).

\begin{figure}[t]
    \centering
    \includegraphics[width=\columnwidth]{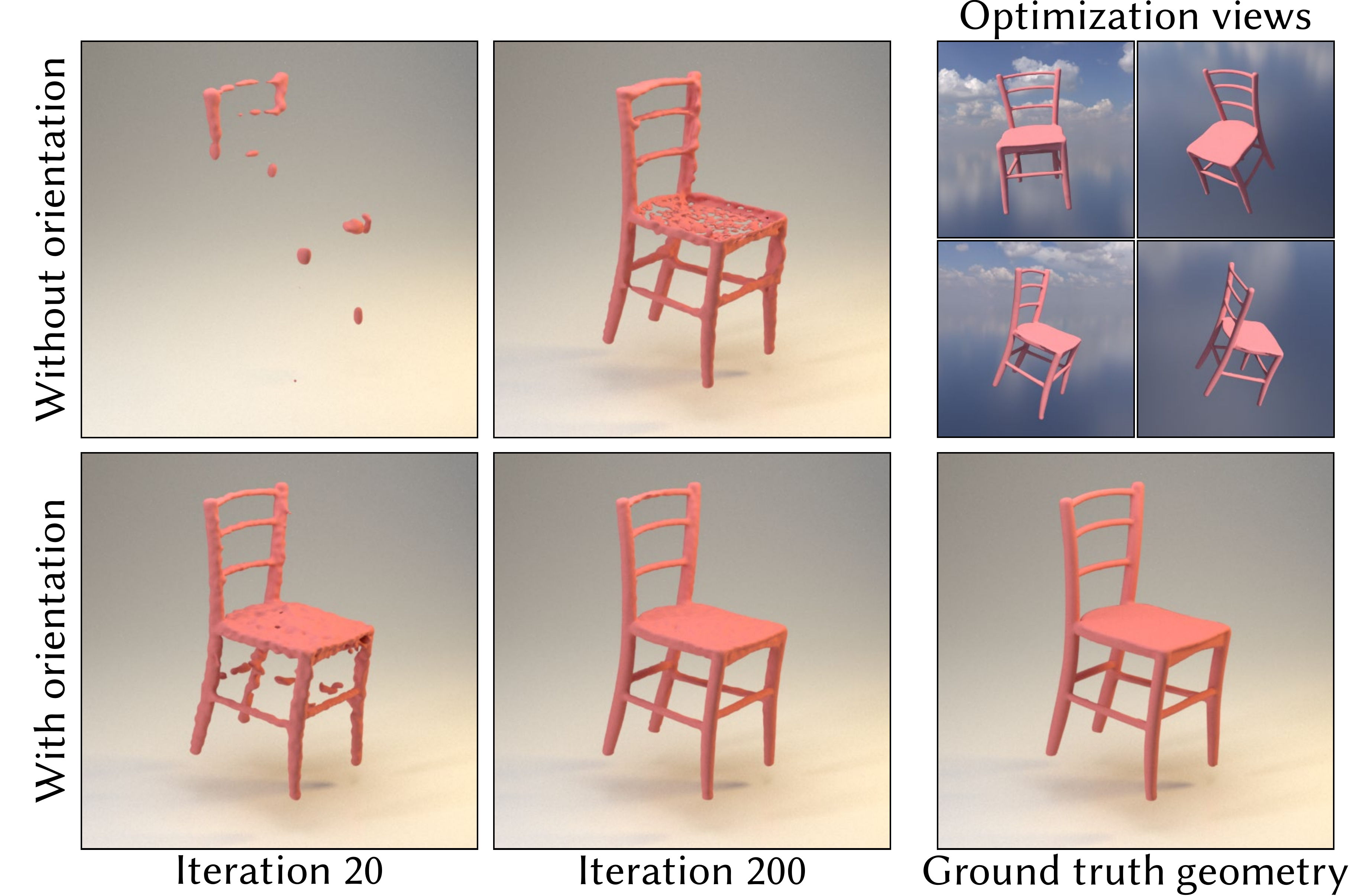}
    \caption{%
        \label{fig:orientation-importance}%
        \textbf{Importance of the orientation field.}
        We compare reconstructions done \emph{with} and \emph{without} an orientation field $\orientation$.
        The top row without $\orientation$ uses an isotropic normal distribution.
        A lack of orientation information dramatically slows down convergence and produces incorrect meshes.}
\end{figure}

Anisotropy is crucial for physically based inverse rendering.
\autoref{fig:orientation-importance} demonstrates this: assuming a uniform normal distribution for surface patches produces incorrect results.
This happens because the reference scene is surface-based, where light reflection is highly anisotropic---a property that isotropic distributions fail to capture.

This concludes our derivation via non-local surface perturbations.
To reinforce these results and gain a deeper understanding:
\begin{enumerate}[leftmargin=6.3mm]
    \item \autoref{sec:appendix_derivation} re-derives \autoref{eqn:manyworlds-deriv} by analyzing standard surface derivatives and extending them into space.
    \item \autoref{sec:expvol} frames our approach using the mathematical language of random volumes.
\end{enumerate}

\subsection{Primal rendering}
\label{sec:manyworld-primal}

Differentiable rendering pipelines normally repeatedly render a primal
image, differentiate a loss, and then backpropagate derivatives.
The previous discussion was only concerned with derivative propagation, and there is thus a question of how to generate a primal image of our extended parameter space\footnote{
    In many applications, derivative propagation arises naturally from automatic differentiation of the primal computation.
    However, this coupling introduces bias in physically based rendering \cite[Section 3.2]{NimierDavid2020Radiative}, which necessitates separate, uncorrelated Monte Carlo simulation of primal rendering and derivative transport.
}.

One option is to only render the surface $\avgsurf$ and discard other potential surfaces.
This approach fails in practice---the extended parameter space disappears to primal rendering, which leads to random optimization behavior in regions distant from the surface.
For example, in a supposedly empty region, occupancy values along the ray will not decay to zero as they may not be involved in the loss computation.

We thus design a primal rendering that incorporates both the surface $\avgsurf$ and the distribution $S$.
Recall that the many-worlds principle manifests in \autoref{eqn:manyworlds-render} as a direct \emph{summation} of non-local perturbations.
This summation is not directly suitable for primal rendering as it yields unbounded values.
Instead, we compute an \emph{average} over all perturbations (not an average over all possible surfaces) by scaling the sum in \autoref{eqn:manyworlds-render} by a factor of $1/s$.
Alternative designs like stochastically rendering individual perturbations per iteration are possible.
Experimentally, we found that scaling by $1/s$ is straightforward, exhibits low variance and promotes rapid convergence.

\paragraph{Relation to ``radiance field loss''}

A recent work \cite{zhang2025radiance} instantiated our many-worlds framework for radiance field reconstruction, specializing it to the setting of pure emission without scattering.
This enables several simplifications:
(1) radiance values are retrieved noise-free from storage (typically a neural network), eliminating the need for BRDF and light integration,
(2) ray marching replaces Monte Carlo sampling, and
(3) rays are traced from fixed pixel centers, so the pixel footprint integral also disappears.

As a result, there is a 1:1 mapping between surface radiance and reference pixel values, allowing individual losses to be defined for each potential surface.
Since radiance contributions from different perturbations are never summed,
their method sidesteps the scaling factor used in our algorithm.
In contrast, this paper targets physically based rendering, where none of these simplifications apply to the harder problem of full global illumination.


\section{Discussion}
\label{sec:discussion}

\paragraph*{Pseudocode}
We present one possible implementation of our many-worlds framework in Mitsuba 3~\cite{jakob2022mitsuba3}.

The following listing uses a variable \texttt{mode} to distinguish between the primal rendering pass and the derivative propagation pass.

\vspace{3mm} 

\begin{minted}[linenos,fontsize=\footnotesize, escapeinside=||, mathescape=true]{python}
mode = "Primal"  # Or "Backward"

def Li(|$\vx$|, |$\vw$|):
    # Pick a segment to interact with a surface patch
    |$k_\mathrm{mw}$| = |$\lfloor k_\mathrm{max}$| * rand()|$\rfloor$|
    return Li_k(|$\vx$|, |$\vw$|, |$k_\mathrm{mw}$|, 0)

def Li_k(|$\vx$|, |$\vw$|, |$k_\mathrm{mw}$|, |$k$|):
    if |$k$| > |$k_\mathrm{max}$|: # Path length exceeds limit
        return 0

    # Radiance estimate from background surface
    s = ray_intersect(|${\avgsurf}$|, |$\vx$|, |$\vw$|)
    |$\vx^\prime$| = |$\vx$| + s * |$\vw$| # Advance to surface
    |$\vw^\prime$|, |$w_\mathrm{brdf}$| = sample_brdf(|$\vx^\prime$|, -|$\vw$|)
    L_bg = Le(|$\vx^\prime$|,-|$\vw$|) + Li_k(|$\vx^\prime$|, |$\vw^\prime$|, |$k_\mathrm{mw}$|, |$k$| + 1) * |$w_\mathrm{brdf}$|
    if |$k$| != |$k_\mathrm{mw}$|: # Segment is before/after the sampled segment
        return L_bg

    # Radiance estimate from sampled surface patch
    t, |$w_\mathrm{surf}$| = sample_surface(rand(), s)
    if mode == "Primal":
        weight = |$w_\mathrm{surf}$| / s # Primal pass: not differentiated
    else:
        weight = |$w_\mathrm{surf}$| # Backward pass: derivative propagation
    |$\vx^\prime$| = |$\vx$| + t * |$\vw$| # Advance to sampled surface
    |$\vw^\prime$|, |$w_\mathrm{brdf}$| = sample_brdf(|$\vx^\prime$|, -|$\vw$|)
    occupancy = |$\alpha$|(|$\vx^\prime$|) # Occupancy at sampled point
    L_fg = Le(|$\vx^\prime$|,-|$\vw$|) + Li_k(|$\vx^\prime$|, |$\vw^\prime$|, |$k_\mathrm{mw}$|, |$k$| + 1) * |$w_\mathrm{brdf}$|
    return lerp(L_bg, L_fg, occupancy) * weight
\end{minted}


\paragraph{Relation to surface derivatives}
Prior work on geometry differentiation have proposed \emph{local derivative formulations}~[\citeauthor{Zhang2020PSDR}~\citeyear{Zhang2020PSDR}, \citeauthor{Zhang2023Projective}~\citeyear{Zhang2023Projective}]
to quantify how small perturbations of a surface affect radiative transport across the entire scene in physical light simulation.

\autoref{sec:appendix_derivation} extends such formulation to measure how tiny changes of \emph{any} hypothetical surface patch within $S$ influence radiative transport.
This offers a quantitative way to re-derive the many-worlds derivative transport law using established theory.
We made the following observations:
\begin{enumerate}[leftmargin=6.3mm]
    \item  Our method \emph{does the right thing} near the surface: the optimization behavior matches surface differentiation algorithms without requiring explicit silhouette sampling.
    \item  Visibility and shading derivatives are combined into a unified expression in the extended parameter space.
    This unification is impossible on the surface $\avgsurf$, as the two derivatives are defined on different domains.
    Unlike prior work---which required separate algorithms to compute these two types of derivatives---our method drastically reduces algorithmic complexity.
\end{enumerate}

\begin{figure}[t]
    \centering
    \includegraphics[width=.8\columnwidth]{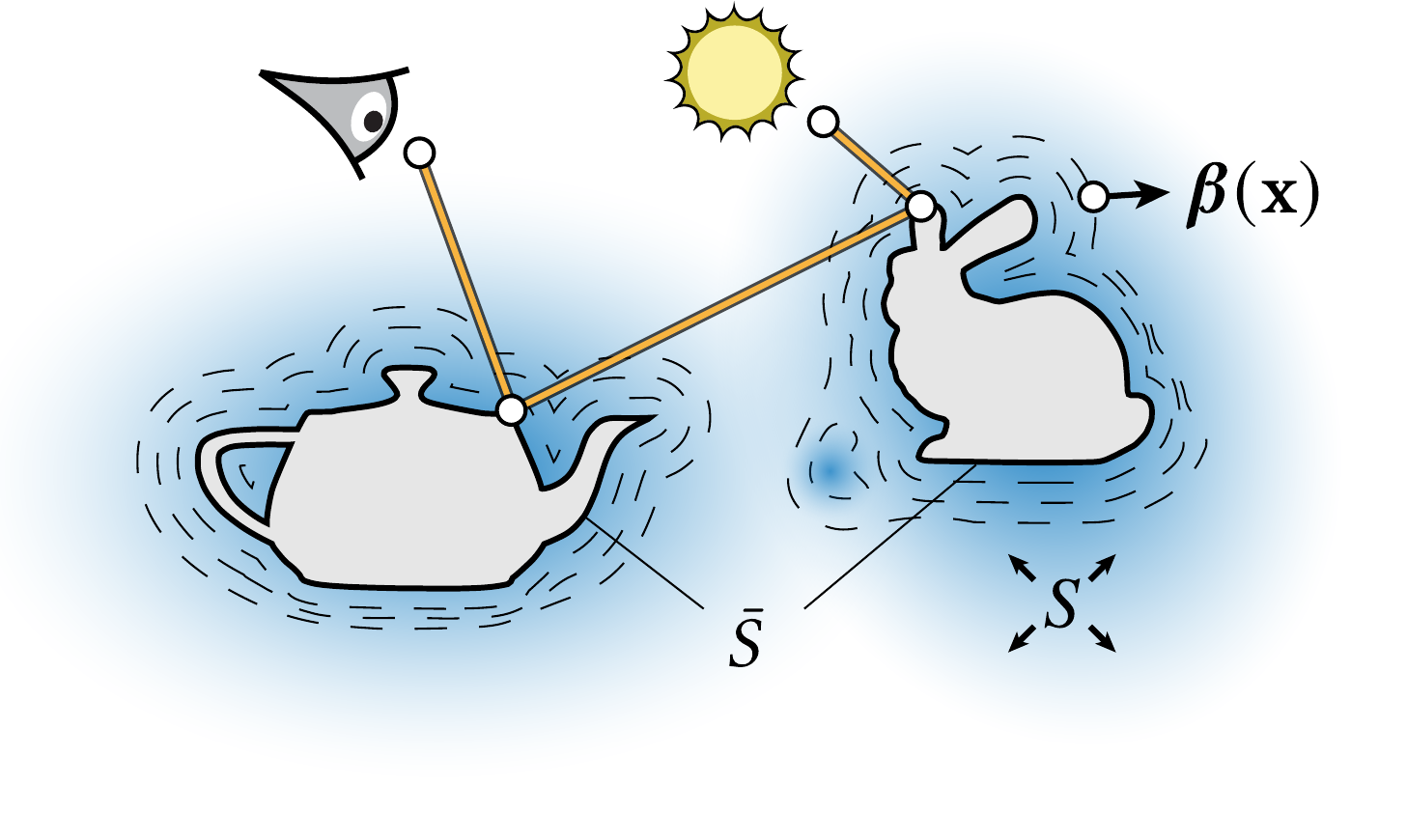}
    \vspace{-9mm}
    \caption{%
        \label{fig:path-and-quantities}%
        Our method refines a surface $\avgsurf$ by optimizing a distribution $S$ of possible surfaces.
        Each hypothetical surface patch drawn from $S$ is independently adjusted to improve the matching between $\avgsurf$ and the target image.
        To define the behavior of possible surfaces in space,
        $S$ is parameterized by an occupancy field $\occupancy(\vx)$ and an orientation field $\orientation(\vx)$.
        }
\end{figure}

\paragraph{Relation to volume rendering}
At a high level, both inverse volume rendering \cite{NimierDavid2022Unbiased} and our method can be interpreted as optimizing \emph{a distribution of surfaces}.
However, the two approaches differ fundamentally in how they interact with this distribution:
when multiple potential surfaces exist along a ray, how do we model their interplay?
\begin{itemize}[leftmargin=6.3mm]
    \item Exponential volume rendering treats interactions as statistically independent events, leading to a memoryless Poisson process where multiple interactions occur, weighted by relative occlusion probabilities \cite{bitterli18framework}.
    \item Our method treats interactions as mutually exclusive events, ensuring potential surfaces along a ray do not interact via shadowing or scattering.
\end{itemize}

Consider a distribution $\surfrv$ modeling a nearly empty scene containing a single infinite plane with an uncertain offset.
Within any realization of this distribution, light traveling towards this plane will scatter \emph{exactly once}.
In contrast, the volume model diffuses the ray-surface interaction into a band of microflakes that cause \emph{arbitrarily long} scattering chains.
This not only leads to a significant increase in computational cost, but also discombobulates the optical interpretation of these interactions:
the effect of multiple scattering must later be approximated as a surface BRDF, a process that in general admits no exact solution and relies on approximation.

Traditional inverse rendering pipeline (which renders the scene to compute a loss) collapses under the mutually exclusive assumption,
because it would render each potential surface as a separate scene, thus unrealistically requiring \emph{every} potential surface to match the reference image.
Instead, we optimize potential surfaces by refining the rendering of a background surface $\avgsurf$,
giving the algorithm a clear convergence target (\autoref{fig:path-and-quantities}).
Such comparative adjustments lead to the notion of \emph{non-local surface perturbations}, and
simultaneous optimization of all perturbations leads to our \emph{many-worlds} derivative transport.


\begin{figure*}[t]
    \centering
    \includegraphics[width=\linewidth]{images/main_result/main_result.pdf}
    \caption{%
        \label{fig:main_recon}%
        \textbf{Multi-view geometry reconstruction}.
        For the \textsc{Deer} scene, all $8$ views are behind the object and we only see the front side in the mirror.
        For \textsc{Polyhedra}, \textsc{Neptune} and \textsc{Fertility}, the object is inside a spherical or cubic smooth glass container.
        The material is known during optimization: the \textsc{Dragon} has a rough gold material, \textsc{Polyhedra} and \textsc{Neptune} are made of copper oxide, and others are diffuse.
        \vspace{1cm}
        }
\end{figure*}


\section{Results}
\label{sec:results}

\begin{figure*}[t]
    \centering
    \includegraphics[width=\linewidth]{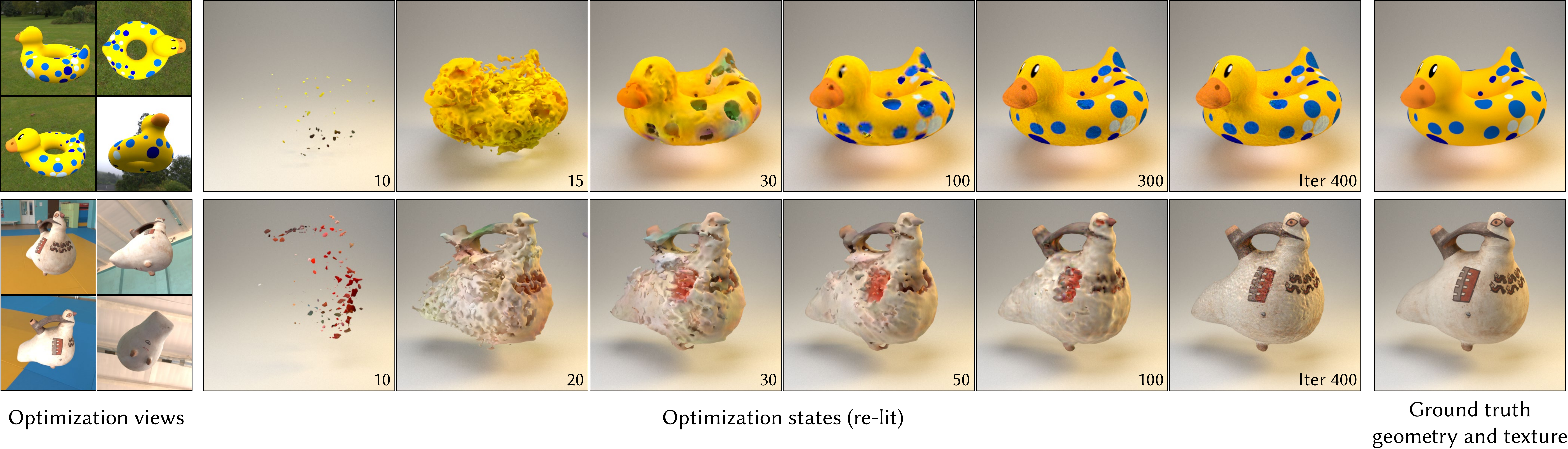}
    \vspace{-7mm}
    \caption{%
        \label{fig:albedo_recon}%
        \textbf{Material optimization}.
        This experiment demonstrates joint optimization of geometry and an albedo texture.
        Our method focuses on geometric optimization, but it is compatible with more general inverse rendering pipelines that furthermore target material and lighting.}
\end{figure*}

\begin{figure*}[t]
    \centering
    \includegraphics[width=0.98\linewidth]{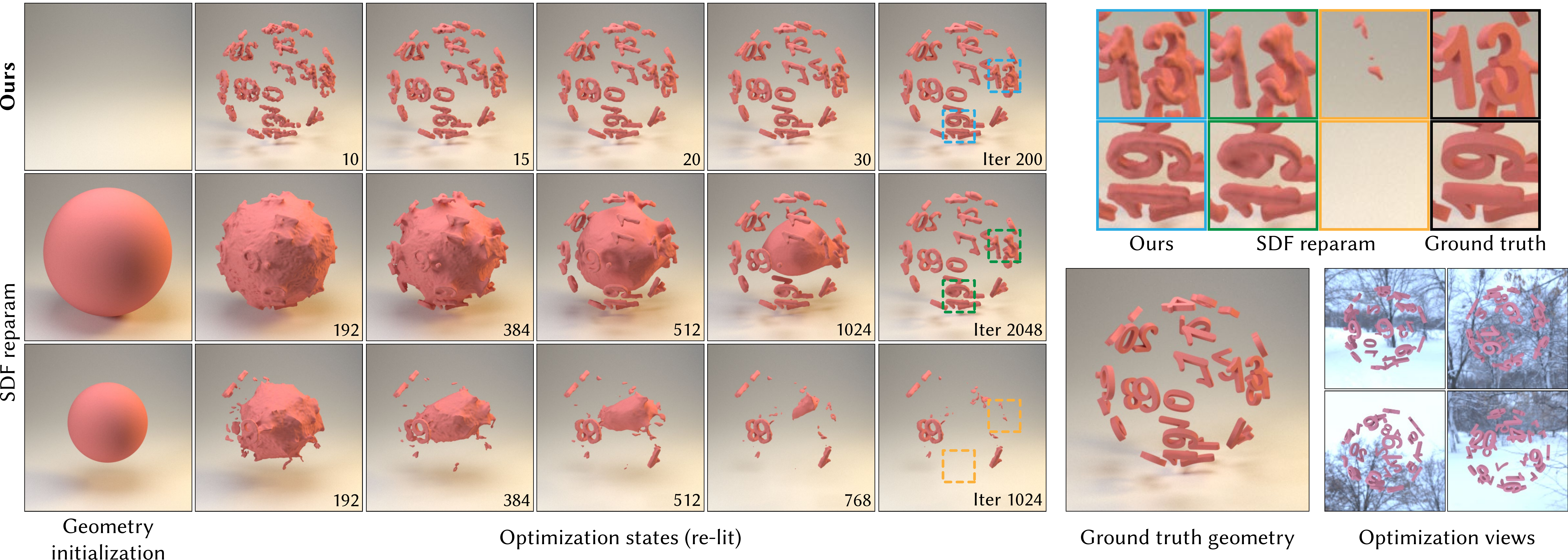}
    \vspace{-2mm}
    \caption{%
        \label{fig:number20}%
        \textbf{Assumption-free optimization.}
        Prior surface optimization methods require an initial guess, which affects the quality of their output.
        We compare our method to an unbiased surface derivative method (SDF reparameterization~\cite{vicini2022differentiable}) using an initialization with a sphere that is either \emph{bigger} (middle) or \emph{smaller} (bottom) than the overall size of the target.
        The experiment shows that the method struggles to reconstruct the target from the smaller sphere, while carving it out of the bigger sphere seems more reliable.
        Our method (top) can optimize without such assumptions, i.e., starting from empty space.}
\end{figure*}

\begin{figure*}[t]
    \centering
    \includegraphics[width=\linewidth]{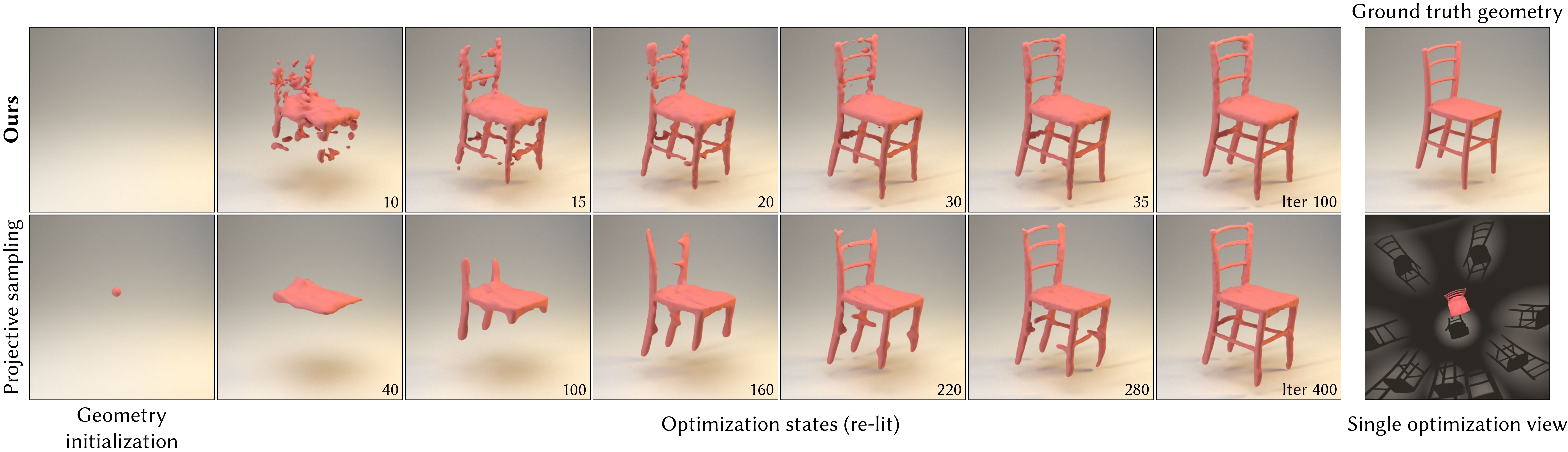}
    \vspace{-6mm}
    \caption{
        \label{fig:shadowchair}%
        \textbf{The benefit of simultaneously optimizing all positions.}
        We replicate an experiment of \citet{Zhang2023Projective} with our method to reconstruct a chair from a single reference image.
        The baseline employs preconditioned gradient descent~\cite{Nicolet2021Large} to locally deform a triangle mesh.
        In each iteration, the pixels rendering the legs propagate gradients to the scene, causing their progressive extrusion in a thin region of overlap between tentative object and the chair in the reference image.
        Derivatives outside of the region of overlap are discarded.
        Our method observes and uses all gradients starting from the very first iteration, enabling faster convergence.}
\end{figure*}

\begin{figure*}[t]
    \centering
    \includegraphics[width=\linewidth]{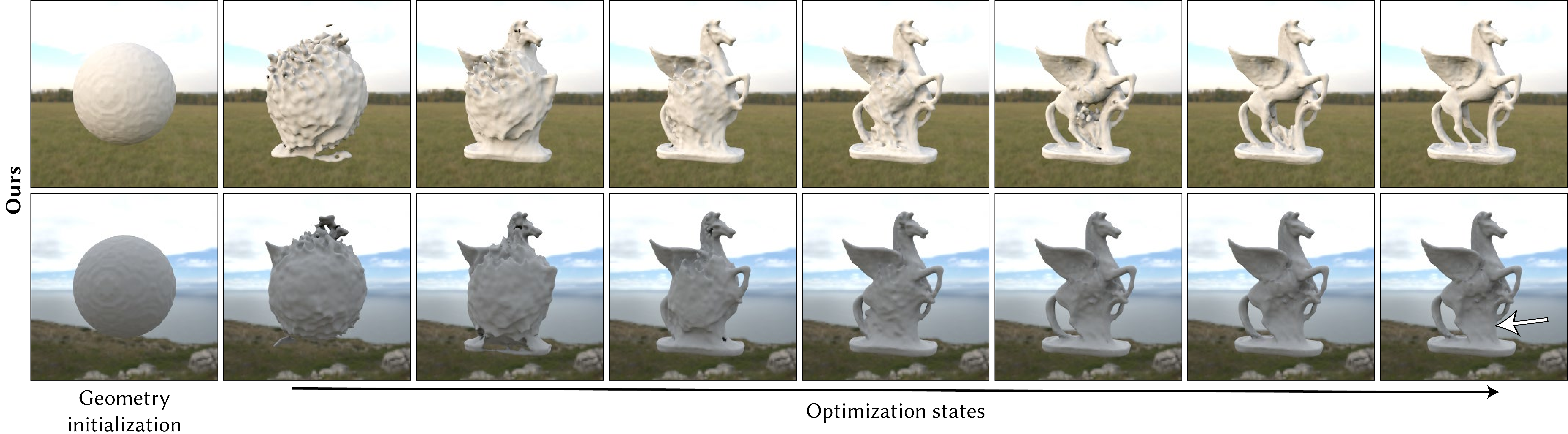}
    \vspace{-5mm}
    \caption{
        \label{fig:hole}%
        \textbf{Limitation on handling interior topological changes.}
        Our method extends surface perturbations only to the exterior and does not accommodate interior topological changes --- such as transforming a sphere into a donut.
        We demonstrate this using a sphere initialization under two lighting conditions.
        While shading derivatives can sometimes produce correct holes (top row), it could fail under alternative lighting conditions (bottom row).}
\end{figure*}

Prior work on physically based inverse rendering often includes comparisons of \emph{forward derivatives} to other competing methods or finite difference-based reference derivatives.
They reveal the change in rendered pixels when perturbing a single scene parameter.
However, such a comparison is neither applicable nor meaningful in our method: first, our method computes the extended derivative on a higher-dimensional domain, which therefore cannot be compared to existing methods.
Second, the need for such comparisons is motivated by the complexity of formulations that deal with discontinuous visibility, but differentiation our representation is ``trivial'' and therefore not interesting.
Because of this, we demonstrate the correctness and performance of our method solely through end-to-end optimizations.

All results presented in this paper exclusively use many-worlds derivatives and \emph{explicitly disable surface derivatives} on $S$, even for the albedo optimization demonstrated in \autoref{fig:albedo_recon}.

\paragraph*{Multi-view reconstructions}
\autoref{fig:main_recon} presents multi-view object reconstructions of a known material in various settings using our method.
All experiments use the Adam optimizer \cite{kingma2014adam}.
We found that disabling momentum in the early iterations helped avoid excessive changes while the occupancy field was still very far from convergence. Alternatively, a stochastic gradient descent optimizer produces comparable results.
During the reconstruction, each view is rendered at a $512\!\times \!512$ resolution.

The last four rows of \autoref{fig:main_recon} show reconstructions involving perfectly specular surfaces.
No prior PBR method could handle such scenes: reparameterization-based methods would need to account for the extra distortion produced by specular interactions, while silhouette segment sampling methods would need to find directional emitters through specular chains.
Both are complex additional requirements that would be difficult to solve in practice, while the problem simply disappears with the many-worlds formulation.

\paragraph*{Albedo reconstruction}
This paper primarily focuses on geometry, but our method can also be used to optimize materials or lighting.
\autoref{fig:albedo_recon} presents experiments where we jointly optimize the geometry and albedo texture of an object.
We store this spatially varying albedo on an additional 3D volume that parameterizes the BSDF of the many-worlds representation.

\paragraph*{Benefits of assumption-free geometry priors}
\autoref{fig:number20} compares the convergence rate of our method to a technique that evolves an SDF using reparameterizations \cite{vicini2022differentiable}.
These experiments demonstrate that our method requires significantly fewer iterations because new surfaces can be materialized from the very first iteration.

We use $20$ random optimization views, with each view rendered at $256\!\times\!256$ pixels.
Both methods utilize a geometry grid resolution of $128^3$.
The time required to perform one iteration of the optimization for one view is $0.25$ seconds for our method, $0.38$ seconds for the large sphere initialization and $0.32$ seconds for the small sphere initialization.
Our method benefits from the efficiency of ray-triangle intersections, whereas the SDF representation relies on a more costly iterative sphere tracing algorithm.
These measurements also show how sphere tracing slows down with smaller steps due to complicated geometry near a ray, as evidenced by the slower performance of the larger sphere initialization.

\paragraph*{Optimizing all positions at once}
\autoref{fig:shadowchair} replicates the chair reconstruction experiment of a recent work by \citet{Zhang2023Projective} to demonstrate the benefits of the extended parameter space.

\paragraph*{Interior topological changes}
\autoref{fig:hole} illustrates a limitation of our method: it does not robustly handle interior topological changes.
This issue arises because many-worlds derivatives extend only to the exterior of $\avgsurf$, leaving the interior unsupervised, much like traditional surface evolution methods.
When attempting to create a hole, we rely on shading derivatives to bend the surface inward, which sometimes produces a hole as desired (top row).
However, optimization like this is sensitive to lighting conditions; as shown in the bottom row, a different lighting can cause the optimization to stall.

To address this limitation,
\citeauthor{mehta2023theory} \shortcite{mehta2023theory} proposed explicitly testing whether creating a cone-shaped hole is beneficial, and \citeauthor{zhang2025radiance} \shortcite{zhang2025radiance} suggested sampling the background surface stochasticity to occasionally permit visibility through high-occupancy regions.
We leave the exploration of these extensions for future work.

The sphere initialization is used solely to demonstrate this limitation; with an assumption-free initialization, our method converges correctly and more quickly in both lighting conditions.

\begin{figure}[t]
    \centering
    \includegraphics[width=1.0\columnwidth]{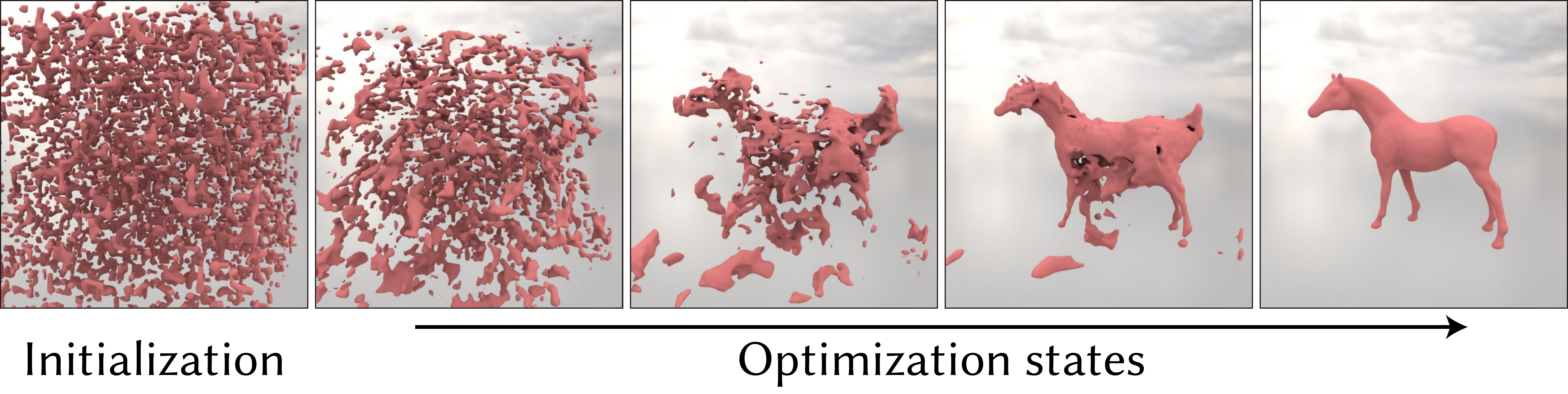}
    \vspace{-6mm}
    \caption{%
        \label{fig:subtractive}%
        \textbf{Subtractive changes.} Many-worlds derivatives match surface
        derivatives when sampled close to the surface. We initialize the scene
        with dense geometry to demonstrate the robustness of our method against
        subtractive changes.}
\end{figure}

\paragraph*{Subtractive changes}
\autoref{fig:subtractive} shows the optimization states for an initialization with random geometry.
We used $16$ optimization views and the same setup as in \autoref{fig:main_recon}.
Since the many-worlds derivative extends the surface derivative domain without approximations, our method naturally address scenarios that surface derivatives can handle---in this case, removing superfluous geometry and deforming the rest to reconstruct the desired object.

\begin{figure}[t]
    \centering
    \includegraphics[width=1.0\columnwidth]{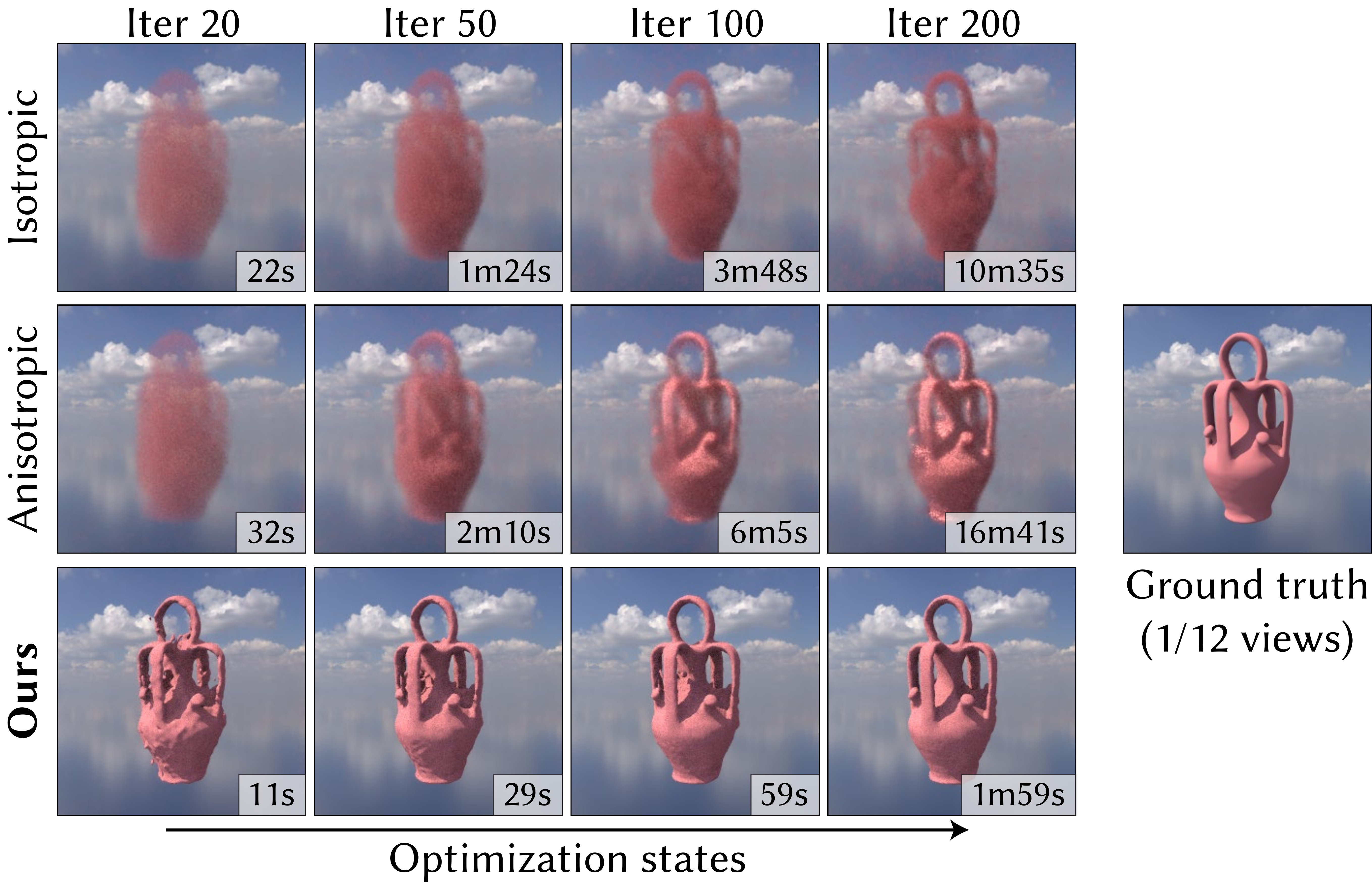}
    \vspace{-6mm}
    \caption{%
        \label{fig:volume_comparison}%
        \textbf{Comparison with volume reconstructions.} We compare
        reconstruction results using exponential volume inversion and the
        proposed many-worlds approach. Isotropic volumes (top) cannot replicate
        the surface appearance, while anisotropic micro-flakes (middle) perform
        somewhat better. However, simulating for interactions between worlds
        (e.g. to model exponential attenuation) introduces substantial
        computational overheads in both volume reconstruction methods.
        Extracting the final BSDF and surface from the volume poses additional
        challenges. Our method (bottom) produces a high-fidelity surface
        reconstruction in a small fraction of the time.}
\end{figure}

\paragraph*{Comparison with volume reconstructions}
\autoref{fig:volume_comparison} presents results and equal-iteration timings comparing our method to volume-based inversion.
For the latter, we set the maximum path depth of the underlying volumetric path tracer to 3 interactions, as larger values significantly degrade performance without improving visual fidelity.
All experiments use the same number of samples per pixel, and each optimization iteration uses all 12 optimization views. The anisotropic volume model uses the SGGX phase function \cite{heitz2015sggx}.
The volume models are initially as fast as ours but slow down significantly as the volume thickens, which is a consequence of iterative steps needed to resolve coupling between different parts of the exponential volume.
Besides reducing speed, this coupling leads to degraded reconstruction quality at equal iteration count.
The many-worlds approach is algorithmically simpler, produces a better result in less time, and directly outputs a mesh with materials that are ready to be relit, without the need for additional optimization to extract a surface BRDF from phase functions.

\paragraph*{Timings.}
\autoref{tab:timing} lists the average computation time per gradient step and view for several scenes.
These timings were measured on an AMD Ryzen~3970X Linux workstation with an NVIDIA RTX~3090 graphics card, which our implementation uses to accelerate ray tracing.
We limited the maximum path depth for every scene to a reasonable value.
For certain scenes, we also disabled gradient estimation for some path segments.
For example, in the shape reconstructions inside a glass object, the first ray segment (from the camera to the glass interface) and the last ray segment cannot intersect the many-worlds representation.


\section{Conclusion and future work}
\label{sec:conclusion}
Correct handling of discontinuous visibility during physically based inversion of geometry has presented a formidable challenge in recent years.
Instead of proposing yet another method to solve this challenging problem, many-worlds inverse rendering shows that there are completely different ways to approach it.
Using a notion of non-local perturbations of a surface, our method successfully synthesizes complex geometries from an initially empty scene.

Our work lays the theoretical foundation and validates this theory with a first implementation.
However, the details of this implementation are still far from optimal and could benefit from various enhancements.
For example, we derive the local orientation from a field that also governs occupancy, which is straightforward but also introduces a difficult-to-optimize nonlinear coupling.
Extending the model with a distribution of orientations would further allow it to consider multiple conflicting \mbox{explanations at every point.}

Reconstructing an object involves a balance between \emph{exploration} to consider alternative explanations in $S$ and \emph{exploitation} to refine parameters of the current explanation $\avgsurf$.
Our method aggressively pursues the exploration phase using a uniform sampling strategy but lacks a mechanism to effectively exploit its knowledge.
We find that it often reconstructs a good approximation of a complex shape in as little as 20 iterations, impossibly fast compared to existing methods, but then requires 500 iterations for the seemingly trivial task of \mbox{smoothing out little kinks.}

Our implementation of the many-worlds derivative is based on a standard physically based path tracer, but previous works on differentiable rendering have shown the benefits of moving derivative computation into a separate phase using a \emph{local formulation}.
This phase starts the Monte Carlo sampling process where derivatives locally emerge (e.g., at edges of a triangle mesh in the case of visibility discontinuities).
Integrating the local form of our model with occupancy-based sampling could refine the background surface with a more targeted optimization of its close neighborhood.

\begin{acks}
    This project has received funding from the European Research Council (ERC) under the European Union's Horizon 2020 research and innovation program (grant agreement No 948846).
\end{acks}

\begin{table}[t]
    \centering
    \caption{%
        \label{tab:timing}%
        We measure the average time needed to optimize one $512^2$ pixel image
        for different scenes. The reported time covers all overheads including
        the primal rendering pass, the uncorrelated derivative propagation pass,
        the optimizer step and scene update.}
    \vspace{-3mm}
    \scalebox{0.9}{    
        \begin{tabular}{lccccc}
            \toprule
                    & \multicolumn{1}{p{1cm}}{\centering Path \\ depth}
                    & \multicolumn{1}{p{0.8cm}}{\centering AD \\ depth}
                    & \multicolumn{1}{c}{\raisebox{-1.2ex}{spp}}
                    & \multicolumn{1}{c}{\raisebox{-1.2ex}{grad spp}}
                    & \multicolumn{1}{c}{\raisebox{-1.2ex}{time (s)}} \\
            \midrule
            \textsc{Heptoroid} & 2 & 2 & 128 & 32 & 0.41 \\
            \textsc{Dragon} & 2 & 2 & 128 & 32 & 0.42 \\
            \textsc{Deer} & 4 & 2 & 128 & 32 & 0.75 \\
            \textsc{Polyhedra} & 4 & 2 & 64 & 32 & 1.10 \\
            \textsc{Neptune} & 5 & 3 & 256 & 32 & 2.56 \\
            \textsc{Fertility} & 5 & 3 & 256 & 32 & 2.06 \\
            \bottomrule
        \end{tabular}
    }
\end{table}


\bibliographystyle{ACM-Reference-Format}
\bibliography{main.bib}

\appendix
\section{Extended surface derivatives}
\label{sec:appendix_derivation}

Our method extends surface optimization by introducing non-local perturbations to a surface $\avgsurf$.
Along each light path, the method tests how potential surfaces \emph{could} exist at sampled positions as perturbations to $\avgsurf$.
By optimizing these hypothetical surfaces, we effectively explore an extended parameter space $S$ to refine $\avgsurf$.

Previous work [\citeauthor{Zhang2020PSDR}~\citeyear{Zhang2020PSDR}, \citeauthor{Zhang2023Projective}~\citeyear{Zhang2023Projective}] established the \emph{local formulation} of surface derivatives, which measures how an infinitesimal surface change affects radiative transport in the entire scene.
We extend their formulation to measure how infinitesimal changes in \emph{any} hypothetical surface patch within $S$ influence radiative transport.
This provides a quantitative way to rederive our method.
In this section, we show that the resulting derivatives match the many-worlds derivative transport.

This analysis leads to two critical results:
\begin{itemize}
\item Many-worlds derivatives reduce to conventional surface derivatives (shading and boundary) when evaluated on $\avgsurf$.
\item We unify shading and boundary derivatives into a single term in the extended domain, enabling an algorithmically simpler implementation.
\end{itemize}

\subsection{Conventional surface derivatives}
In physically based rendering, surface differentiation involves two types of derivatives: the \textit{shading derivative} and the \textit{boundary derivative}.
The shading derivative applies to the entire visible surface for a given viewpoint, primarily affecting appearance through the shading normal and local material properties.
In contrast, the boundary derivative (corresponding to ``occlusion'' in \autoref{fig:teaser}) is defined only along the visibility silhouette curve as seen from a viewpoint.
It adjusts the object's visibility contour and is the main driver of shape optimization.

Both derivatives can be expressed as integrals over path space, a high-dimensional domain encompassing all possible light paths.
To simplify our derivation, we write them in the \emph{three-point form} (\cite[Section 8.1]{VeachThesis}) that isolates the derivative contribution from a single ray segment.
While the integral remains high-dimensional, this segment-specific form encapsulates most of the dimensions within the incident radiance term $L_i$ and the incident importance term $W_i$, making the formulation more tractable.

\subsubsection{Shading derivative}
\begin{figure}[t]
    \centering
    \includegraphics[width=0.48\textwidth]{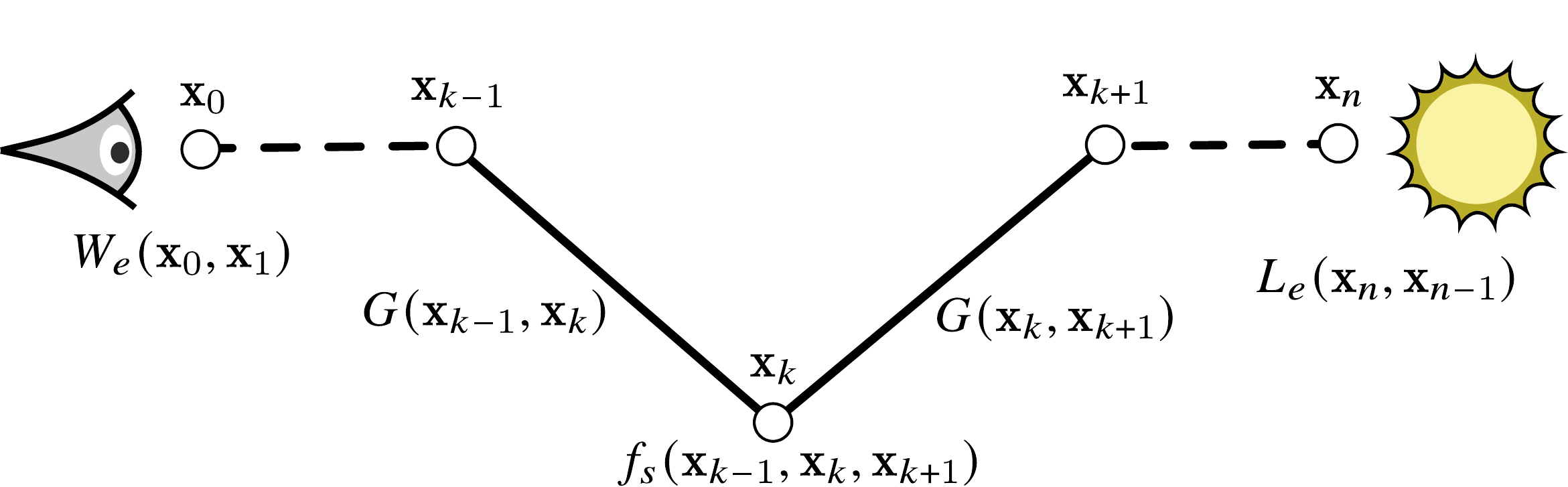}
    \caption{A light path with $n$ segments connecting the camera to the emitter.}
    \label{fig:lightpath}
\end{figure}
Consider the radiance contribution of a $n$-segment light path as illustrated in
\autoref{fig:lightpath}. The measurement contribution function is given by:
\begin{align}
  f(\bar{\x}) =
    W_e(\x_0,\, & \x_1) G(\x_0, \x_1) \\
    \notag
    & \bigg[
    \prod_{i=1}^{n-1} f_s(\x_{i-1}, \x_i, \x_{i+1})  G(\x_{i}, \x_{i+1})
    \bigg]
    L_e(\x_{n}, \x_{n-1}),
\end{align}
where $f_s$ is the BSDF function, $G$ is the standard geometric term including visibility $\mathcal{V}$, and $L_e$ and $W_e$ refer to the emitted radiance and importance.
This equation contributes to a measurement---for instance an image pixel color $I$---following the path space integral \cite{VeachThesis}:
\begin{align*}
    I = \int_\Omega f(\bar{\x}) \dmu,
\end{align*}
where $\mathrm{d}\mu = \prod_{i=0}^{n} \dA(\x_i)$ is the area product element.

Without loss of generality, we assume that the emitted radiance $L_e$ and the
camera model do not depend on the scene parameter $\theta$ (i.e., detached).
Differentiation of the contribution function with respect to $\theta$ yields:
\begin{align}
    \frac{\partial f(\bar{\x})}{\partial\theta} =
    \sum_{k=1}^{n-1}
    W_i(&\xkm, \xk) G(\xkm, \xk)
    \\ \notag
    & \partial_\theta
    \Big[
    f_s(\xkm, \xk, \xkp) G(\xk, \xkp)
    \Big]
    L_i(\x_{k+1}, \x_{k}),
\end{align}
where functions $L_i$ and $W_i$ denote incident radiance and importance:
\begin{align*}
    & L_i(\x_{k+1}, \x_{k}) = L_e(\x_{n}, \x_{n-1}) \Pi_{i=k+1}^{n-1}
    [G(\x_{i}, \x_{i+1}) f_s(\x_{i-1}, \x_i, \x_{i+1}) ],  \\
    & W_i(\x_{k-1}, \x_k) = W_e(\x_0, \x_1) \Pi_{i=1}^{k-1} [G(\x_{i-1}, \x_i)
    f_s(\x_{i-1}, \x_i, \x_{i+1})].
\end{align*}
This gives the three-point form shading derivative of $I$:
\begin{align}
  \label{eq:segment_shading_derivative}
  \bigg[ \frac{\partial I}{\partial\theta} \bigg]_s
    & = \int_\Omega \partial_\theta f(\bar{\x}) \dmu
    \notag \\
    & =
        \int_{\mathcal{M}\times \mathcal{M}}
        W_i(\x_{k-1}, \x_k) \, G(\x_{k-1}, \x_k)
    \notag \\
    &
        \underbrace{
        \int_{\mathcal{M}}
        \partial_\theta
        \Big[
        G(\xk, \xkp)
        f_s(\xkm, \xk, \xkp)
        \Big]
        L_i(\xkp, \xk) ~\dA(\xkp)
        }_{\partial_\theta \widehat{L_o}(\xk, \,\xkm)}
    \notag \\
    &
        \quad \dA(\xk)\dA(\xkm).
\end{align}
The term in the curly bracket is abbreviated as $\partial_\theta \widehat{L_o}(\xk, \xkm)$,
which captures the derivative
stemming from only the \emph{current interaction} at $\xk$.
The derivatives from later interactions along the path are incorporated in their respective three-point form integrals.

\subsubsection{Boundary derivative}

\begin{figure}[t]
    \centering
    \includegraphics[width=0.35\textwidth]{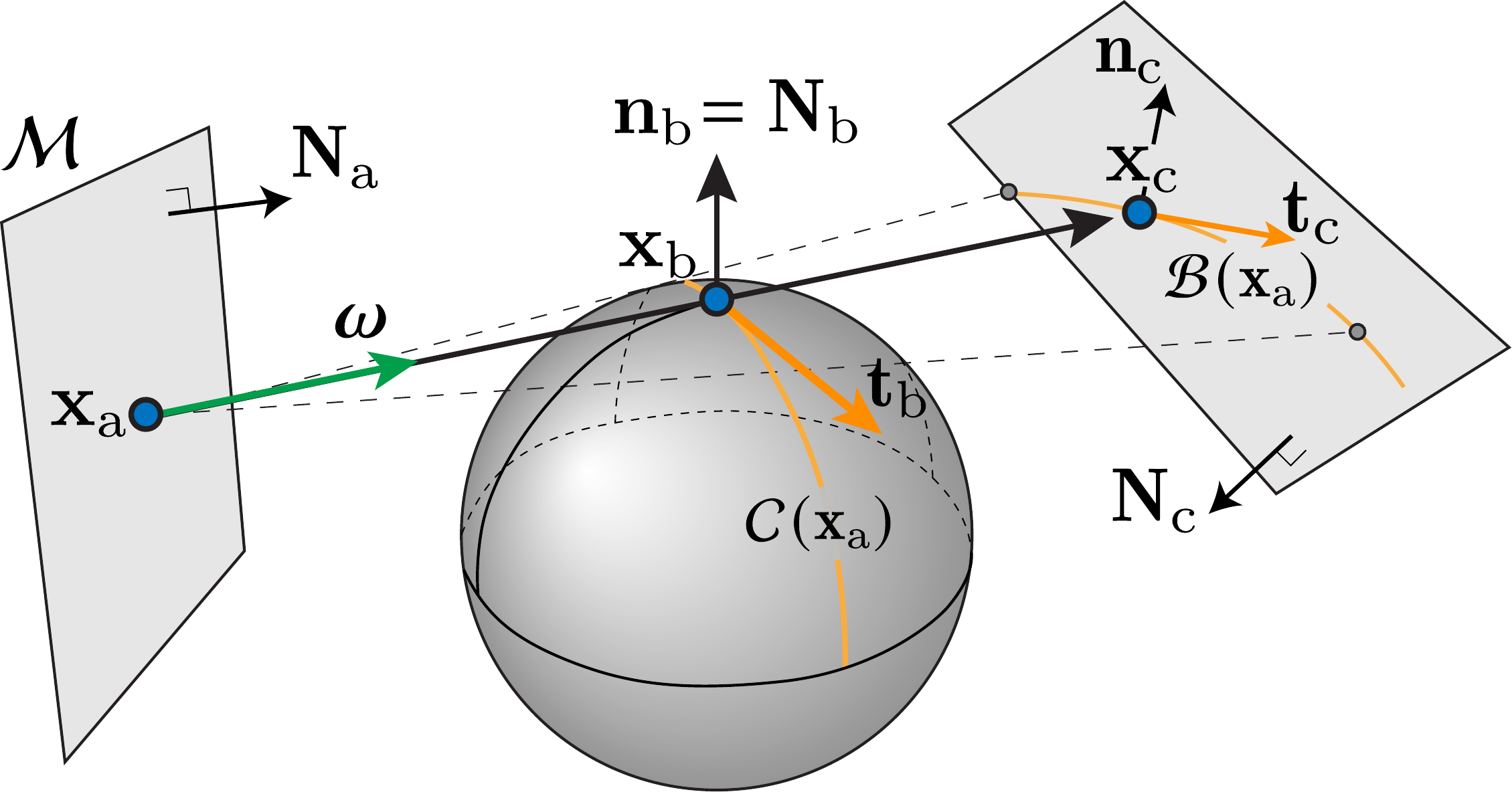}
    \caption{\textbf{Surface boundary derivative}.
    The boundary derivative contains a motion term that tracks how fast the boundary segment $(\xa, \xc)$ moves along the normal direction $\nb$,
    which is orthogonal to the viewing direction $\bomega$ and the silhouette curve direction $\tb$.}
    \label{fig:silhouette_old}
\end{figure}

We start from Equation (43) in \citeauthor{Zhang2020PSDR}'s work \shortcite{Zhang2020PSDR} to derive the segment-specific boundary derivative integral.
The derivation is similar to the one shown in the supplementary material of \cite{Zhang2023Projective} with a different objective---we aim to express the integral such that it is local to the boundary segment, rather than to the silhouette point.
One can alternatively derive the same result starting from Equation (4) in \cite{Zhang2023Projective}.

The original boundary derivative integral, without reparameterization, states that (see \autoref{fig:silhouette_old} for illustration):
\begin{align}
    \bigg[ \frac{\partial I}{\partial\theta} \bigg]_b =
    \int_\mathcal{M}\int_{\mathcal{B}(\xa)}
    L_d(\xb, \xa)\,
    & G(\xa,\xc) \,
    W_i(\xa, \xc)  \\
    \notag
    & (\partial_\theta \xc\cdot \nc)
    ~ \dl(\xc) \dA(\xa),
\end{align}
where $(\xa, \xc)$ form a boundary segment that makes contact with the boundary point $\xb$, and the domain $\mathcal{B}(\xa)$ is the boundary curve on the $\xc$ side surface as seen from $\xa$.
The function $L_d$ defines radiance difference between the foreground and the background as $L_d(\xb, \xa)=L_o(\xb, \bomega) - L_i(\xb, -\bomega)$.
The inner product measures the motion of $\xc$ along its in-surface normal direction $\nc$.

This integral is already specific to a ray segment, but the motion term $\partial_\theta \xc \cdot \nc$ is a derived quantity from $\partial_\theta \xb \cdot \nb$.
They are related as follows \cite[Supplementary Eq. (9)]{Zhang2023Projective}:
\begin{align*}
    \frac{\partial_\theta \xc\cdot \nc}
    {\partial_\theta \xb\cdot \nb}
    =\frac{\lengthac}{\lengthab\|\nb\times\Nc\|}.
\end{align*}
We also change the length measure $\dl(\xc)$ to $\dl(\xb)$ \cite[Supplementary Eq. (3)]{Zhang2023Projective}:
\begin{align*}
    \frac
    {\dl(\xc)}
    {\dl(\xb)}
    =
    \frac
    {\lengthac\|\bomega\times\tb\|}
    {\lengthab\|\bomega\times\tc\|}.
\end{align*}
Futhermore, we use the following identity (\cite[Supplementary Eq. (10)]{Zhang2023Projective}):
\begin{align*}
    \|\bomega\times\tc\| \, \|\nb\times\Nc\|
    =
    |\bomega\cdot\Nc|.
\end{align*}
Combining these equations, we obtain:
\begin{align}
    \bigg[ \frac{\partial I}{\partial\theta} \bigg]_b =
    \int_\mathcal{M} \int_{\mathcal{C}(\xa)}
    L_d(\xb, \xa)\,
    \frac{| \bomega \cdot \Na | \, \| \bomega \times \tb \|}{\lengthab^2} \,
    & W_i(\xa, \xb)  \\
    \notag
    (\partial_\theta \xb\cdot \nb)
    ~ & \dl(\xb) \dA(\xa),
\end{align}
where the boundary domain $\mathcal{C}(\xa)$ is the visibility silhouette curve on the occluder as seen from $\xa$.
Rewriting this result by converting the length measure from scene space to hemisphere space, we get:
\begin{flalign}
    \label{eq:segment_boundary_derivative}
    \quad\,
    \bigg[ \frac{\partial I}{\partial\theta} \bigg]_b =
    \int_{\mathcal{M}} \int_{\mathcal{C}(\xa)}
    L_d(\xb, -\bomega) \, & W_i(\xa, \bomega) \, |\bomega\cdot\Na|
    \\ \notag
    & (\partial_\theta \bomega \cdot \nb)
    ~\dl(\bomega) \dA(\xa).
    &&
\end{flalign}
Intuitively, the motion term $\partial_\theta \bomega \cdot \nb$ measures how rapidly the occluder moves in the direction perpendicular to the viewing ray.
This motion is weighted by the radiance difference between the foreground and background, and the derivative arising from this motion is transported to the sensor akin to regular radiance \cite[Section 3.1]{NimierDavid2020Radiative}.

\subsection{Many-worlds derivatives}
We now extend surface derivatives into the extended domain to also quantify how
\emph{potential surface patches} along a ray segment affect the radiative transport.

\subsubsection{Shading derivative}
\begin{figure}[t]
    \centering
    \includegraphics[width=0.28\textwidth]{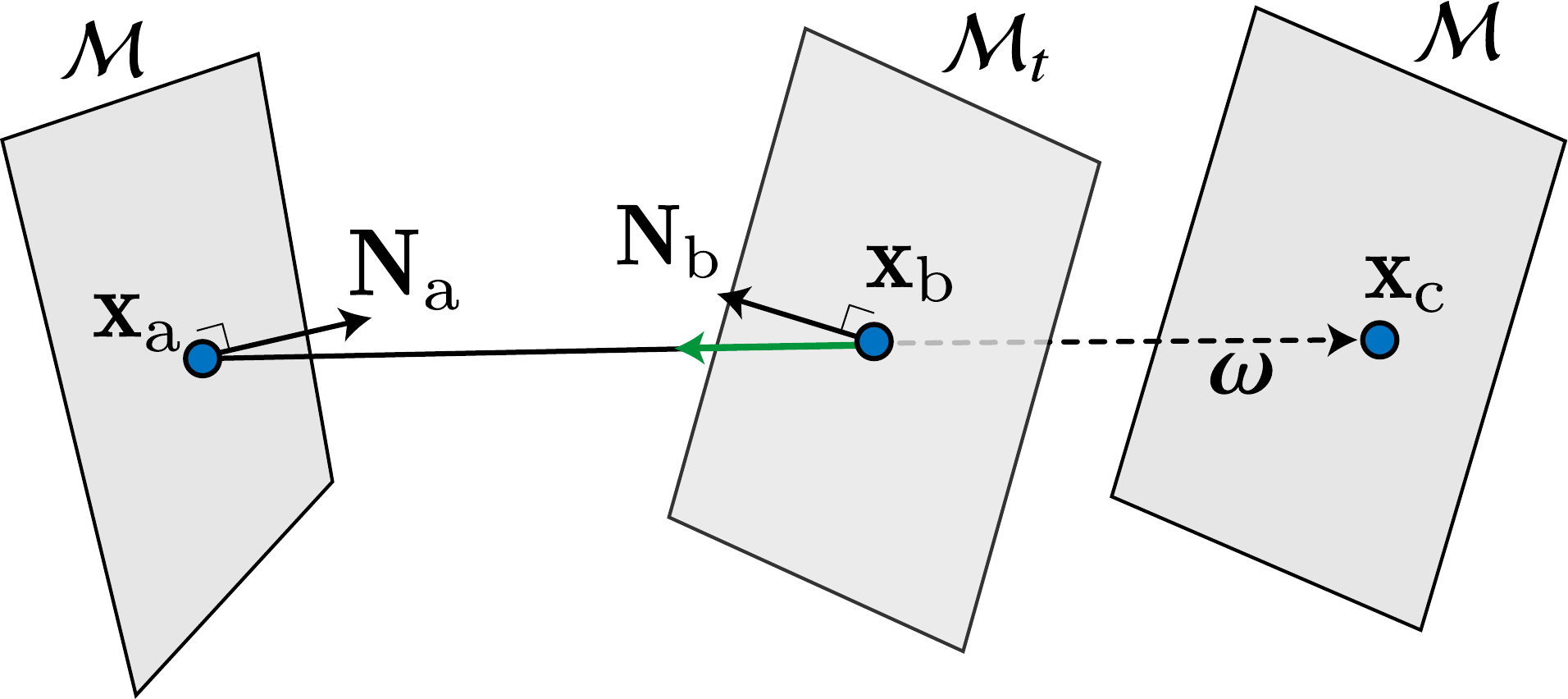}
    \vspace{-2mm}
    \caption{\textbf{Shading derivatives in many-worlds transport}. Along a ray
        segment $(\xa, \xc)$, we consider potential interactions with a hypothetical surface at $\xb$ as a perturbation to the background surface.
        The shading derivatives from all such interactions along the ray segment are accumulated.}
    \label{fig:shading_deriv}
\end{figure}
We write the shading derivative as an integral over potential surface patches along the ray segment $(\xa, \xc)$, weighted by the probability of each patch's existence:
\begin{align}
    \label{eq:many_worlds_shading_derivative}
        \int_0^{\lengthac} \, & \alpha(t) \, \bigg[ \frac{\partial I}{\partial\theta} \bigg]_s ~\dt
        \overset{\eqref{eq:segment_shading_derivative}}{=}
        \int_0^{\lengthac}  \int_{\mathcal{M}\times \mathcal{M}_t}
    \\ \notag
        &
            \alpha(t) \,
            \Big[
            W_i(\xa, \xb) \, G(\xa, \xb) \, \partial_\theta \widehat{L_o}(\xb, \xa)
            \Big]
        ~\dA(\xb)   \dA(\xa)  \dt,
\end{align}
where $\xb$ is on a hypothetical surface $\mathcal{M}_t$ (\autoref{fig:shading_deriv}).
Note that the ray direction may not be perpendicular to the surface normal
$\dA(\xb) \dt = \dV(\xb) / |\bomega\cdot \Nb|$, allowing us to write the integral as:
\begin{flalign}
    \quad\,\,
        = & \int_{\mathcal{R}^3} \int_{\mathcal{M}} \alpha(\xb) \, W_i(\xa, \xb)
    \\ \notag
        &
        \quad\quad
         \frac{|\bomega\cdot \Na| \mathcal{V}(\xa \leftrightarrow \xb)}{\lengthab^2} \,
        \partial_\theta \widehat{L_o}(\xb, \xa)
        ~\dA(\xa) \dV(\xb). &&
\end{flalign}
By chaging from an area measure to a solid angle measure, we can cancel
the remaining geometric term and absorb the visibility function $\mathcal{V}$
to obtain:
\begin{flalign}
    \label{eq:many_worlds_shading_derivative_final}
    \quad\,\,
        = \int_{\mathcal{R}^3} \int_{S^2}
        \alpha(\xb) \, W_o(\xb, -\bomega) \, \partial_\theta \widehat{L_o}(\xb, -\bomega)
        ~\dw \dV(\xb). &&
\end{flalign}
This derivative has a simple form from which we can identify the following
properties:
\begin{itemize}
    \item It is local to the position $\xb$ where the interaction happens.
    \item The shading derivative originating from this interaction is weighted
    by the probability of the surface's existence and is uniformly radiated
    in all directions to be received by the sensor.
\end{itemize}

\subsubsection{Boundary derivative}
A challenge in analyzing the boundary derivative for hypothetical surfaces is that existing theory from prior work only analyzes \emph{visibility silhouettes} on occluders.
This becomes problematic as a potential surface patch is not always on such a silhouette:
its surface normal $\Nb$ may not be orthogonal to the viewing direction $\bomega$.
Such a patch still influences radiative transport as it occludes radiance traveling from the background.
However, this type of occlusion does not occur in local surface evolution, where occlusion is restricted to the visibility silhouette.

We therefore introduce an extended form of the boundary derivative whose value matches the conventional boundary derivative on a visibility silhouette curve, but is also well-defined for any surface.

\begin{figure}[t]
    \centering
    \includegraphics[width=0.4\textwidth]{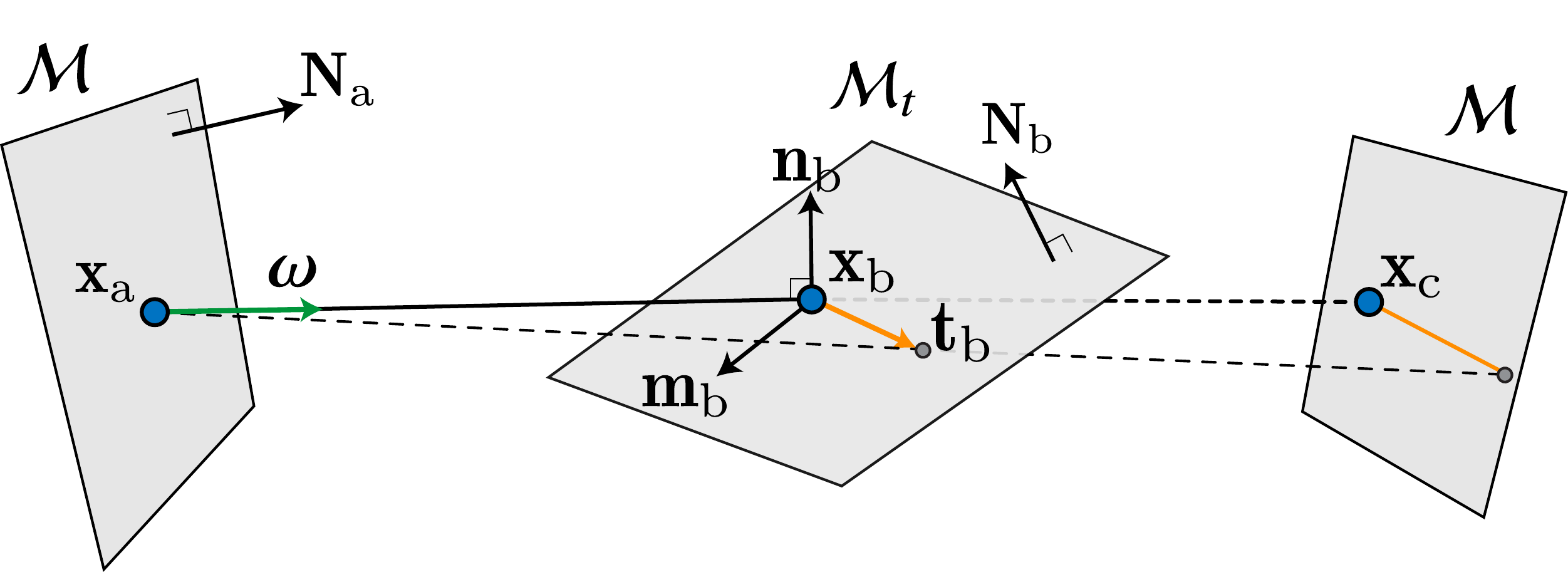}
    \caption{ \textbf{Extended form of the surface boundary derivative}.
    For conventional boundary derivatives, the silhouette direction $\tb$
    is only defined on the visibility silhouette.
    In the extended case, we define it for any surface, allowing silhouette occlusion to be interpreted as probabilistic surface existence.}
    \label{fig:silhouette_gen}
\end{figure}

Does such an extension make sense? Intuitively, the boundary derivative examines the color difference between the foreground and the background, and the motion term determines if more of the foreground or the background should be visible.
This concept closely mirrors the derivative of \textit{occupancy}, where adjusting the probability of the surface existence decides the visibility balance between the surface and the background.
From this perspective, the ``boundary'' derivative logically extends to hypothetical surfaces with non-tangential ray intersections.
Although ``boundary'' may be a misnomer in these cases, we retain the term for consistency with existing literature.

\paragraph*{Extended form of surface boundary derivatives}
For any surface located at $\xb$ with normal $\Nb$, we define the \textit{extended silhouette direction} $\tb$ as the in-surface direction along which the dot product $\bomega\cdot \Nb$ remains constant.
The conventional silhouette direction is a special case where $\bomega\cdot \Nb=0$.

Let $\nb$ be the viewing direction normal, which is orthogonal to both the viewing direction $\bomega$ and the extended silhouette direction $\tb$ (\autoref{fig:silhouette_gen}).
In the extended case, it is important to distinguish between the surface normal $\Nb$ and the viewing direction normal $\nb$.
This subtle difference is not present in the conventional case, as $\Nb=\nb$ on the visibility silhouette.

We denote the local motion at $\xb$ as
\begin{align}
    \label{eq:motion_abb}
    \vth(\xb) := \partial_\theta \xb \cdot \Nb
\end{align}
since it quantifies the rate at which occlusion changes at the interaction point.

Let $\mb$ be a unit vector in the surface such that $\{\tb, \Nb, \mb\}$ forms an orthonormal basis.
the motion relates to ray direction $\bomega$ as \cite[Supplementary Eq. (9)]{Zhang2023Projective}:
\begin{align}
    \label{eq:normal_motion_local}
    \partial_\theta \bomega \cdot \nb =
    \frac{\|\nb\times \mb\|}{\lengthab} \, \partial_\theta \xb \cdot \Nb.
\end{align}

The length measure $\dl(\bomega)$ is in the solid angle domain, and it relates to the length measure $\dl(\xb)$ on the surface as
\begin{align}
    \label{eq:2d_length_measure}
    \dl(\bomega) = \frac{\|\bomega\times \tb\|}{\lengthab} \dl(\xb).
\end{align}

Using these equations, the extended boundary derivative from \autoref{eq:segment_boundary_derivative} can be rewritten as:
\begin{align}
    \label{eq:gen_segment_boundary_derivative}
        \bigg[ \frac{\partial I}{\partial\theta} \bigg]_b
        & \overset{\eqref{eq:segment_boundary_derivative}}{=}
        \int_{\mathcal{M}} \int_{\mathcal{C}(\xa)}
        L_d(\xb, -\bomega) \, W_i(\xa, \bomega) \, |\bomega\cdot\Na|
    \notag \\
        & \quad\quad\quad\quad\quad\quad\quad\quad\quad\quad\quad\quad\quad  
        (\partial_\theta \bomega \cdot \nb)
        ~\dl(\bomega) \dA(\xa)
    \notag \\
        & \overset{\!\!\!\!(\ref{eq:motion_abb},\ref{eq:normal_motion_local})\!\!\!\!}{=}
        \int_{\mathcal{M}} \int_{\mathcal{C}(\xa)}
        L_d(\xb, -\bomega) \, W_i(\xa, \bomega) \, |\bomega\cdot\Na|
    \notag \\
        & \quad\quad\quad\quad\quad\quad\quad\quad\quad\,\,\,\,\,
        \frac{\|\nb\times \mb\|}{\lengthab} \, \vth(\xb)
        ~\dl(\bomega) \dA(\xa)
    \notag \\
        & \overset{\eqref{eq:2d_length_measure}}{=}
        \int_{\mathcal{M}} \int_{\mathcal{C}(\xa)}
        L_d(\xb, -\bomega) \, W_i(\xa, \bomega) \, |\bomega\cdot\Na|
    \notag \\
        & \quad\quad\quad\quad
        \frac{\|\nb\times \mb\| \, \|\bomega\times \tb\|}{\lengthab ^ 2} \, \vth(\xb)
        ~\dl(\xb) \dA(\xa).
\end{align}

\paragraph*{Boundary derivative in many-worlds transport}
At each interaction position $\xb$, let $\dl(\xb)$ be the length measure along the extended silhouette direction $\tb$
and $\dn(\xb)$ be the length measure along the surface normal $\Nb$.
Together, they define an area element $\dA(\xb)=\dl(\xb) \dn(\xb)$ which quantifies how this patch occludes the background.

Analogous to shading derivatives, we write the many-worlds boundary derivative of $I$ as an integral over all possible surface interactions along the ray segment $(\xa,\xc)$:
\begin{align}
    \label{eq:many_worlds_boundary_derivative}
        \iint  \,  \bigg[ \frac{\partial I}{\partial\theta} \bigg]_b & ~ \dn \dt
        \overset{\eqref{eq:gen_segment_boundary_derivative}}{=}
        \int_0^{\lengthac}  \int_{\mathcal{M} \times \mathcal{M}_t}
            L_d(\xb, -\bomega) \, W_i(\xa, \bomega) \,
    \\ \notag
        &
            |\bomega\cdot\Na| \,
            \frac{\|\nb\times \mb\| \, \|\bomega\times \tb\|}{\lengthab ^ 2} \, \vth(\xb)
        ~ \dA(\xb)  \dA(\xa)  \dt,
\end{align}
The basis $\{\tb, \Nb, \bomega\}$ leads to a volume element
\begin{align}
    \label{eq:volume_measure}
    \dV(\xb) = |\bomega\cdot \mb| ~\dl(\xb)\dn(\xb)\dt.
\end{align}
Note that $\mb$, $\nb$ and $\Nb$ are co-planar since they are all orthogonal to $\tb$.
We therefore obtain an identity \cite[Supplementary Eq. (10)]{Zhang2023Projective}:
\begin{align}
    \label{eq:cos_sin_sin_identity}
    \|\bomega\times \tb\| \, \|\nb \times \mb\|=|\bomega \cdot \mb|.
\end{align}
Using these equations, we simplify the boundary derivative as:
\begin{flalign}
        \eqref{eq:many_worlds_boundary_derivative}
        \overset{\eqref{eq:volume_measure}}{=}
        & \int_{\mathcal{M}} \int_{\mathcal{R}^3}
        W_i(\xa, \bomega) \,  L_d(\xb, -\bomega) \,
        \frac{|\bomega\cdot \Na|}{\lengthab^2}
    \notag \\
        & \quad\quad\quad\quad\quad\quad
        \frac{\|\bomega\times \tb\|\, \|\nb \times \mb\|}{|\bomega \cdot \mb|}
        \, \vth(\xb)
        ~ \dV(\xb) \dA(\xa)
    \notag \\
    \label{eq:many_worlds_boundary_derivative_before_motion}
        \overset{\eqref{eq:cos_sin_sin_identity}}{=}
        & \int_{\mathcal{R}^3} \int_{\mathcal{S}^2}
        W_o(\xb, -\bomega) \,  L_d(\xb, -\bomega) \,
        \vth(\xb)
        ~ \dw \dV(\xb).
\end{flalign}

We define the local motion $\vth(\xb)$ within its canonical space:
\begin{align}
    \label{eq:many_worlds_motion}
    \vth(\xb) = \partial_\theta \alpha(\xb).
\end{align}
This definition contrasts with prior work that computes the surface boundary derivative using an implicit field representation.
In those works, the motion is defined in the scene space, resulting in the following Jacobian determinant \cite{stam2011velocity}:
\begin{equation}
    \label{eq:2d_normal_velocity}
    \partial_\theta \p \cdot \mathbf{N} = \frac{\partial_\theta \alpha(\p)}{\|\nabla \alpha(\p)\|},
\end{equation}
that relates the normal velocity of a surface point $\p$ to the derivative of its implicit representation.
This Jacobian determinant is not required since our method optimize the surface patch directly by modifying its occupancy value, rather than locally deforming it.
The scaling factor in \autoref{eq:2d_normal_velocity} can bias the many-worlds derivative by erroneously considering neighboring occupancy values.
This is more apparent when considering the fact that the gradient norm $\|\nabla \alpha\|$ could be infinitely large near the mean surface or the opposite, it could drop to zero in the case of an initialization where all positions share the same occupancy value.

This yields the final form of the boundary derivative in many-worlds transport:
\begin{flalign}
    \label{eq:many_worlds_boundary_derivative_final}
    \eqref{eq:many_worlds_boundary_derivative_before_motion}
    \overset{\eqref{eq:many_worlds_motion}}{=}
    & \int_{\mathcal{R}^3} \int_{\mathcal{S}^2}
    W_o(\xb, -\bomega) \,  L_d(\xb, -\bomega) \,
    \partial_\theta \alpha(\xb)
    ~ \dw \dV(\xb).
&&
\end{flalign}

\subsubsection{Many-worlds derivative}
Combining \autoref{eq:many_worlds_shading_derivative_final} and \autoref{eq:many_worlds_boundary_derivative_final},
we derive a derivative integral that captures the overall impact of non-local surface perturbations:
\begin{align}
        \label{eq:many_worlds_derivative}
        \frac{\partial I}{\partial\theta}
        & =
        \int_{\mathcal{R}^3} \int_{\mathcal{S}^2}
        W_o(\xb, -\bomega) \,
        \Big[
            \alpha(\xb) \, \partial_\theta \widehat{L_o}(\xb, -\bomega) \, +
    \notag \\
        & \quad\quad\quad\quad\quad\quad\quad\quad\quad\quad
            \partial_\theta \alpha(\xb) \, L_d(\xb, -\bomega)
        \Big]
        ~ \dw \dV(\xb)
    \notag \\
        & =
        \int_{\mathcal{R}^3} \int_{\mathcal{S}^2}
            W_o(\xb, \bomega) \,
            \partial_\theta
            \Big[
                L_d(\xb, \bomega) \, \alpha(\xb)
            \Big]
        ~ \dw \dV(\xb),
\end{align}
where the background surface in the radiance difference function $L_d$ is \emph{detached}.
\autoref{eq:many_worlds_derivative} represents the final form of the many-worlds derivative, expressed in an alternative domain but equivalent to \autoref{eqn:manyworlds-deriv}.

This derivation suggests that our approach is an extension of standard surface evolution methods.
Conventional techniques evolve surfaces by locally modifying shading properties or adjusting geometry to change occlusion relationships,
with such perturbations restricted to the surface itself.
Instead, we generalize this concept by analyzing hypothetical surface patches in space, leading to the same well-defined shading and occlusion-based perturbations in an extended domain.

A key challenge in this extension is that any position along a ray can now contribute to surface evolution, rather than being restricted to a single intersection point.
To address this, we adopt the ``many-worlds'' perspective: treating all possible interactions as mutually exclusive events.
This ensures that a desired change is uniformly applied across all potential interactions.


\section{A random volume perspective}
\label{sec:expvol}

\subsection{Transport in an exponential random volume}
\label{sec:expvol-overview}
We begin by reviewing light transport in an exponential random volume as described by the radiative transfer equation.
This equation gives the incident radiance $\Li$ along a ray $(\vx,\vw)$, accounting for radiative gains (in-scattering, emission) and losses (out-scattering, absorption) along the segment reaching up to the nearest ray-surface intersection at \mbox{$s=\inf\{s'\mid\vx+s'\vw\in \surfaces\}$} or \mbox{$s=\infty$} if none exists:
\begin{equation}
    \label{eqn:rte}
    \Li(0)=\int_0^s T(t)\left[\mus(t)\, \Ls(t) + \mua(t)\, \Le(t)\right]\mathrm{d}t +T(s)\, \Lo(s).
\end{equation}
The first term models contributions of the volume, while the second accounts for a potential surface at $t=s$, whose outgoing radiance $\Lo(s)$ is attenuated by the medium.

The functions $\mua(t)$ and $\mus(t)$ specify the volume's \emph{absorption} and \emph{scattering} coefficient, whose sum gives the \emph{extinction} \mbox{$\mut=\mua+\mus$}.
The in-scattered radiance
\begin{equation}
    \label{eqn:inscat}
    \Ls(t)=\int_{\mathcal{S}^2}L_i(t, \vw')\,f_p(t,\vw,\vw')\dw'
\end{equation}
is the product integral of incident radiance and the \mbox{\emph{phase function} $f_p$}.
Finally, the \emph{transmittance}
\begin{equation}
    \label{eqn:transmittance}
    T(t)=\exp\left(-\int_0^t\mu_t(t') ~\mathrm{d}t'\right)
\end{equation}
ties everything together: it establishes the connection to particle distributions and ensures energy conservation by accounting for self-shadowing (or more generally, self-extinction).

\subsection{Many-worlds transport}
\label{sec:tenuous-vol}
Building on the discussion in \autoref{Sec:motivation}, we aim to prevent nonsensical scattering and shadowing between multiple potential surfaces along a ray.
This can be achieved by modeling the interaction with a \emph{tenuous} volume, whose density is diluted by a factor of $\varepsilon$, which reduces the scattering and extinction coefficients $\mus$ and $\mut$.
This low amount of extinction ensures at most one interaction with the volume along a ray segment, leading to the following transmittance approximation:
\begin{equation}
    \label{eqn:transmittance-linear}
    T(t)\approx 1-\int_0^t\mu_t(t')\,\mathrm{d}t',
\end{equation}
which follows from $e^{-x}\approx 1 - x + O(x^2)$.
Plugging this into the RTE~\eqref{eqn:rte} and discarding a second-order term ($\mus\mut\sim\varepsilon^2$) yields the approximation $\Li(0)\approx\Lie(0)$, where the latter is defined as
\begin{align}
    \label{eqn:tenuous-vol}
    \Lie(0)\!\coloneqq\!\!\!\int_0^s\!\!\!\left[\mus(t)\, \Ls(t)+\mua(t)\,\Le(t)-\mut(t)\,\Lo(s)\right]\mathrm{d}t +
    L_o(s),\!
\end{align}
In other words, a tenuous volume is governed by a linearized RTE, where higher-order terms vanish.
The superscript $\varepsilon$ refers to quantities with a dilution factor $\varepsilon$.

To complete this model, we must instill meaning into the terms $\mut$, $\mus$, and $f_p$.
Prior work often did so using volumetric analogs of microfacet surface models known as \emph{microflakes}~\cite{Jakob2010Microflake}.
Applications include optimization, volumetric level of detail, energy-conserving random walks, and neural fields~\cite{heitz2015sggx,heitz2016multiple,vicini2021non,Loubet2018Microflake,Zhang2023NeMF}.

It is worth noting that the theory of microflakes actually builds on a more general notion of \emph{anisotropic radiative transport}~\cite{shultis1988radiative,Jakob2010Microflake}, which adopts directionally varying extinction and scattering coefficients arising from a random distribution of oriented particles:
\begin{align}
    \mut(\vx,\vw)&=\rho(\vx) \int_{S^2}\sigma(\vw,\vw')\,D(\vx, \vw') \dd\vw'\\
    \mus(\vx,\vw)&=\rho(\vx) \int_{S^2}a(\vx, \vw, \vw')\,\sigma(\vw,\vw')\,D(\vx, \vw') \dd\vw',
\end{align}
where $\rho(\vx)$ is the particles' number density at $\vx$, $D(\vw)$ models the density of their directional orientations, $\sigma(\vw, \vw')$ gives the cross-sectional area of a single particle with orientation $\vw'$ observed from direction $\vw$, and $a(\vx,\vw,\vw')$ models the particles' scattering albedo ($\in[0,1]$).
Microflake theory can then be derived from these expressions by setting $\sigma(\vw,\vw')=\sigma \,|\vw\cdot\vw'|$ to model the projected area of a facet with surface area $\sigma$, and by constructing a phase function $f_p$ based on the \mbox{principle of specular reflection from such a flake.}

However, we do not model a directional distribution $D(\vw)$ in this work.
Instead, we adopt a far simpler particle model that associates \emph{a single} particle orientation $\orientation(\vx)$ with every point $\vx$ (\autoref{fig:alpha_beta_illustration}).

Since there is only one orientation at $\vx$, the distribution $D$ collapses to a Dirac delta function: $D(\vx, \vw)=\delta(\vw-\orientation(\vx))$, which in turn reduces the extinction to a simple product of number density and cross-sectional area
\begin{equation}
    \mut(\vx,\vw)= \rho(\vx)\, \sigma(\vw, \orientation(\vx)).
\end{equation}
Here, we do not specifically model the split into a separate number density and cross-section.
Instead, we define an extinction function that directly evaluates their product:
\begin{align}
    \label{eqn:nonreciprocal-occupancy}
    \mut(\vx,\vw)&=\begin{cases}
        \varepsilon\, o(\vx),&\mathrm{if }\ \vw\cdot\orientation(\vx)>0,\\
        0,&\mathrm{otherwise},
    \end{cases}
\end{align}
where $o(\vx)$ is the \emph{occupancy}, i.e., the point-wise discrete probability that $\vx$ is inside $S$ (\autoref{eqn:occupancy}).
The branch condition encodes an important \emph{nonreciprocal}\footnote{This behavior is non-reciprocal because when the medium interacts along $\vw$ (i.e., if \mbox{$\mu_t(\vx,\vw)>0$}), then the opposite direction is non-interacting (i.e., $\mu_t(\vx,-\vw)=0$).} behavior: light interacting with a back-facing surface presents a nonsensical case, and this term masks those parts of the volume.

The extinction $\mut$ models the \emph{stopping power} of the volume.
For example, the flakes of a microflake volume obstruct light to a lesser degree when it propagates nearly parallel to them, and this manifests via a foreshortening term $|\vw\cdot\vw'|$ in the definition of $\mut$.
This is important to obtain a sensible (energy-conserving, reciprocal) model because these particles mutually interact.
On the other hand, when focusing on a single world in a many-worlds representation, light stops with probability 1 when it encounters a surface regardless of how it is oriented (except for mentioned back-facing case).
This is why our model does not include a similar foreshortening term.
The term $\alpha(\vx)$ models the discrete probability of the world within the ensemble.

We define the \emph{phase function} $f_p$ of the volume as
\[
    f_p(\vx,\vw,\vw')=\frac{f_s(\vx,\vw,\vw')\,|\orientation(\vx)\cdot\vw'|}{a(\vx,\vw,\orientation(\vx))},
\]
which wraps $f_s$, the \emph{bidirectional scattering distribution function} (BSDF) of the many-worlds surface at the point $\vx$.
This definition follows the standard convention that the phase function integrates to $1$, with absorption handled by other terms.
The albedo $a$ provides the necessary normalization constant:
\begin{equation}
    a(\vx,\vw,\orientation(\vx))=\int_{S^2}f_s(\vx,\vw,\vw')\,|\orientation(\vx)\cdot\vw'|\dw'.
\end{equation}
With these definitions, $\mus$ reduces to the \mbox{product of extinction and $a$:}
\begin{align}
    \mus(\vx,\vw)&=a(\vx, \vw, \orientation(\vx))\,\mut(\vx,\vw).
\end{align}
The product of this scattering coefficient and in-scattered radiance $\Ls$ (\autoref{eqn:inscat}) can be seen to compute an extinction-weighted integral of the many-worlds BSDF over projected solid angles:
\begin{align*}
    \mus(\vx,\vw)\Ls(\vx,\vw)=\mut(\vx,\vw)\!\!\int_{S^2}\!\!\Li(\vx,\vw')\,f_s(\vx,\vw,\vw')\,|\beta(\vx)\cdot\vw'|\dw'.
\end{align*}
We further use the sum of the above expression with the absorption-weighted emission to define an extended outgoing radiance function $\Lo$ for points $\vx$ that lie within the many-worlds representation:
\begin{align}
    \mus(\vx,\vw)\Ls(\vx,\vw)+\mua(\vx,\vw)\Lo(\vx,\vw)&\eqqcolon\mut(\vx,\vw)\Lo(\vx,\vw).
\end{align}
Substituting all of these expressions into \autoref{eqn:tenuous-vol}, we obtain
\begin{align}
    \label{eqn:manyworlds-li-perturb}
    \Lie(0)=
    \varepsilon\int_0^s \occupancy(t)\left[\Lo(t)-\Lo(s)\right]\mathrm{d}t +
    L_o(s),
\end{align}
where $\alpha$ gives the occupancy at $t$ and is zero in the back-facing (\mbox{$\orientation(t)\cdot\vw<0$}) case.

This equation quantifies how the many-worlds representation modifies light transport along individual path segments.
For multi-segment paths, we do not model multiple interactions, which would introduce second-order derivative terms beyond our scope.

While \autoref{eqn:manyworlds-li-perturb} captures infinitesimal volumetric effects, our many-worlds framework treats each perturbation as a standalone alternative world (\autoref{Sec:motivation}).
Removing the $\varepsilon$ scaling reveals the \emph{unit perturbation} effect:
\begin{equation}
    \label{eqn:manyworlds-li-perturb-rescaled}
    \Li(0)
    =\int_0^s \occupancy(t)\left[\Lo(t)-\Lo(s)\right]\dd t + L_o(s).
\end{equation}
Under differentiation this gives the many-worlds derivative transport in \autoref{eqn:manyworlds-deriv}.


\section{Technical Details}
\label{sec:implementation}

\subsection{Stochastic surface model}
\label{sec:surface-distr}
Our method evolves a surface $\avgsurf$ by propagating gradients to a stochastic surface model $S$.
We outline one possible parameterization of a stochastic surface model here~\cite{williams2006gaussian,sellan2022stochastic,sellan2023neural,Miller2023Theory,seyb2024microfacets}, noting that this is not our contribution and other formulations can be used.
Our work focuses on the theoretical foundation for optimizing \emph{a distribution of surfaces} without relying on exponential volumes (i.e., ray-surface interactions are mutually exclusive events, rather than statistically independent events), which is largely independent of the specific model used.

An implicit representation $\implicit(\vx)$ of a shape determines whether a position $\vx$ is inside ($\implicit(\vx)<0)$, outside ($\implicit(\vx)>0)$, or on the boundary ($\implicit(\vx)=0)$ of a solid.
This geometric classification is independent of optical properties---for example, points $\vx$ within a refractive material also count as \emph{inside} ($\implicit(\vx)<0$).

In our case, the scene models a distribution of surfaces, which turns $\implicit(\vx)$ into a random variable.
The occupancy $\occupancy(\vx)$ then gives the probability of $\vx$ being on or inside an object:
\begin{equation}
    \occupancy(\vx) = \Pr\{\implicit(\vx) \le 0\}.
\end{equation}

We model $\implicit(x)$ as a \emph{Gaussian process} (GP). Pointwise evaluations of the implicit function are normally distributed:
\begin{equation}
    \implicit(\vx)\sim \mathcal{N}(\mu(\vx), \sigma^2(\vx)),
\end{equation}
which yields an explicit form of the occupancy:
\begin{equation}
    \occupancy(\vx)=\frac{1}{2}\left[1-\erf\left(\frac{\mu(\vx)}{\sqrt{2\sigma^2(\vx)}}\right)\right].
\end{equation}
We assume a constant variance $\sigma^2(\vx)=\sigma^2$ for all $\vx$ in this work.

We also use this representation to assign an orientation to every point based on the expected gradient of the implicit function, i.e.,
\begin{align*}
    \label{eqn:beta-field}
    \orientation(\vx)
    =\frac{E\left[\nabla\implicit(\vx)\right]}{\left\|E\left[\nabla\implicit(\vx)\right]\right\|}
    =\frac{\nabla\mu(\vx)}{\left\|\nabla\mu(\vx)\right\|}.
\end{align*}
Modeling orientation has a crucial impact on the robustness and speed of optimizations~(\autoref{fig:orientation-importance}).

A GP also has the property that evaluations $\implicit(\vx_1), \implicit(\vx_2), \ldots$ follow a joint multivariate normal distribution.
Their auto-correlation is often described using a \emph{kernel} that depends on distance, e.g.:
\begin{equation}
    \corr(\Phi(\vx), \Phi(\vy))=\exp\left(-\gamma\|\vx-\vy\|^2\right)
\end{equation}
where $\corr()$ refers to Pearson's correlation coefficient.
This ensures that nearby points become increasingly correlated, making sudden jumps in realizations of $\Phi(\vx)$ unlikely.

Autocorrelation would be a crucial property if our method involved steps such as sampling concrete surface realizations from $\surfrv$, or if we consider interaction with potential surfaces at multiple different locations, requiring careful modeling of the correlation between them.
However, our method does not depend on such steps.
Its sole interaction with $\surfrv$ is through pointwise evaluations of probabilities, and the extraction of the \emph{background surface} $\avgsurf$:
\begin{equation}
    \avgsurf
    = \{\vx\in\mathbb{R}^3\mid\mu(\vx)=0\}
    = \{\vx\in\mathbb{R}^3\mid\occupancy(\vx)=\nicefrac{1}{2}\}.
\end{equation}
Spatial correlation is therefore not a detail that must be modeled in our implementation of the algorithm.
Other variants of the many-worlds algorithm, for instance those that involve a more general directional distribution, require more parameterization of the GP.

\begin{figure}[t]
    \centering
    \includegraphics[width=0.97\columnwidth]{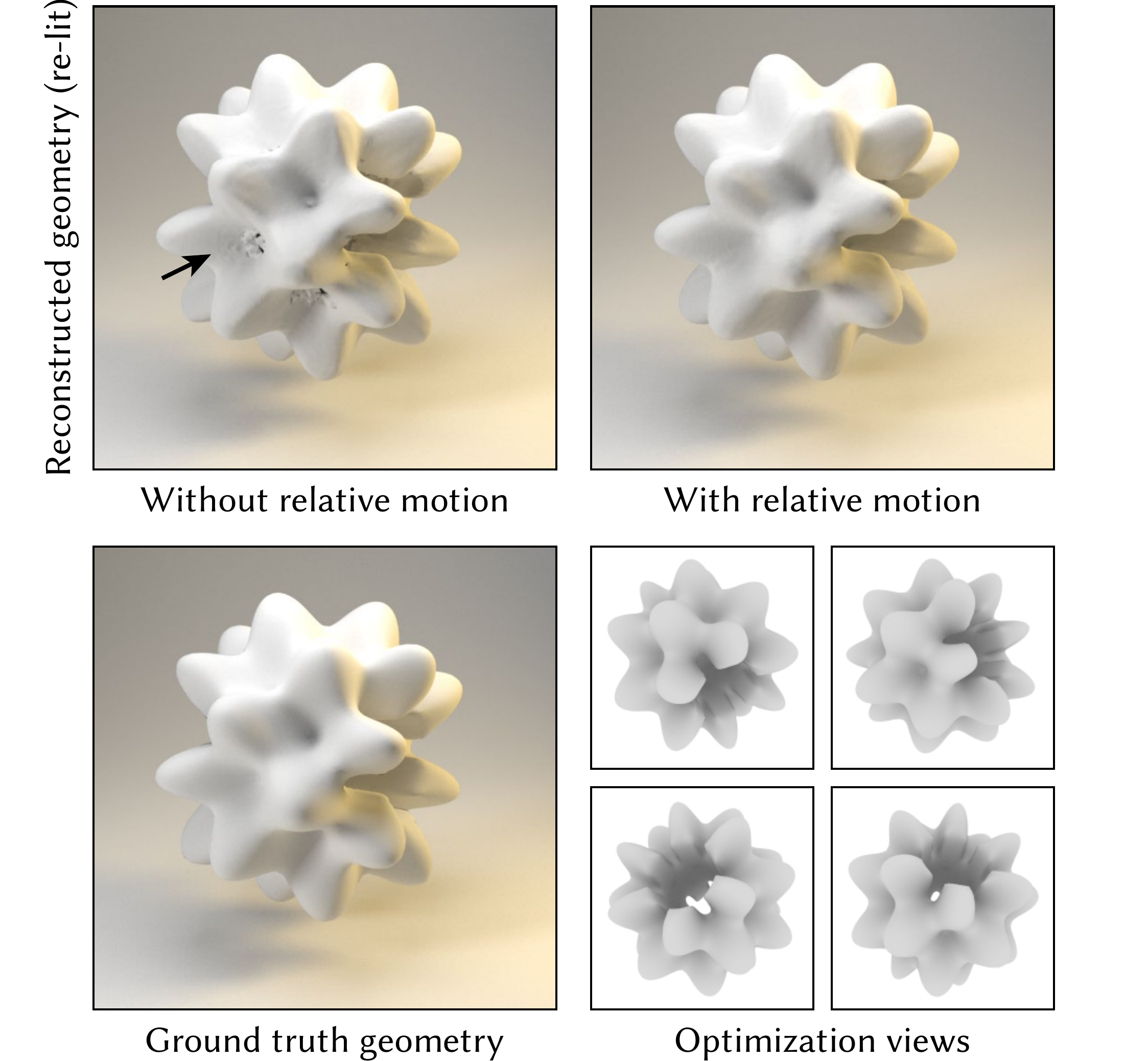}
    \caption{%
        \label{fig:relative_motion}%
        \textbf{Derivative due to relative motion.}
        A diffuse, bumpy torus is rendered under uniform environmental lighting with direct illumination.
        The shading derivative is zero everywhere since the lighting is uniform.
        To reconstruct the concave regions of the torus, we propagate derivatives to the shading point as a result of the relative motion of the shadow ray.}
\end{figure}

\subsection{Implementation details}

\paragraph*{Occupancy and background surface}
We query $\mu(\vx)$ from a grid-based texture using cubic interpolants in our implementation.
Alternatively, a neural representation \cite{mueller2022instant} could be employed for greater flexibility and resolution.

To simulate interactions with the detached background surface $\avgsurf$, we use Marching Cubes~\cite{Lorenson1987Marching} to extract a triangle mesh from $\mu(\vx)$, enabling the use of efficient ray-triangle intersection routines.
Any other isosurface extraction algorithm could in principle be used, since the extraction step and the extracted surface do not need to be differentiable.
Our pipeline does not involve sphere tracing, delta tracking or other iterative algorithms.

\paragraph*{Derivative due to relative motion}
The analysis in \autoref{sec:appendix_derivation} focuses on derivatives caused by direct perturbations to sampled positions along light paths.
However, when differentiating a geometry in a path tracer, indirect shading introduces a unique derivative that arises from the relative motion of the boundary segment. These derivatives are occlusion-induced but are not located on the point of occulsion.

To illustrate, consider the ambient occlusion-style rendering in \autoref{fig:relative_motion}. Here, concave regions (e.g., the bumpy torus's inner valley) never form visible silhouettes from any viewpoint.
The color at any shading point depends solely on whether sampled shadow rays are occluded; in this case, the color depends only on self-occlusion.
While optimizing visibility contours can reconstruct extruded regions, concave regions receive no gradients---their darker appearance stems from self-occlusion, not normal orientation in this scene.

To recover these regions, the system must recognize geometric equivalences: moving the valley floor inward reduces incoming radiance equivalently to moving the valley's ridge outward.
Both operations lead to a motion of the boundary segment, which in turn affects the shading color.

This is not a new concept. Reparameterization-based methods account for this by attaching ray origins in nested reparameterization, while silhouette segment sampling methods propagate boundary derivatives not only to the boundary point but also to the shading point.
As shown in \autoref{sec:appendix_derivation}, our many-worlds approach extends surface derivatives into space while preserving their values on the surface.
However, this extension does not apply to the relative motion derivative because it originates from the fixed background surface $\avgsurf$ rather than from the point of interaction.
One possible solution is to model the background surface as stochastic, similar to recent work on radiance field reconstruction~\cite{zhang2025radiance}.
We leave these extensions for future work.
In this experiment, we instead propose a simple approximation to this derivative by propagating boundary derivatives also to the shading point.
In practice, the relative motion derivative rarely affects optimization, except when the shading derivative is nearly zero---which is why we illustrate this concept using a uniform environmental lighting in \autoref{fig:relative_motion}.

\paragraph{Alternative pamameterization}
The derivation in \autoref{sec:method} discusses the propagation of derivatives to some parameterization of the extended parameter space.
Our choice of parameterization is simple but unnecessarily restrictive.

For instance, it should be possible to construct versions of this framework to model $\orientation$ as a normal distribution rather than a single normal direction.
This could accelerate convergence by allowing the optimizer to simultaneously explore a wider range of surface orientations.
Upon surface extraction, the normal distribution could be interpreted as micro-scale surface roughness, which we leave for future work.

\paragraph{Sampling strategy}

A detail that must be addressed is how \autoref{eqn:manyworlds-render} should be sampled in a Monte Carlo renderer.
Since we don't know the value of the radiance, a natural choice is to sample other terms proportional to their known contribution---in this case, drawing positions proportional to $\occupancy(t)$.
While this is a sensible choice for primal rendering, it leads to a chicken-and-egg problem during optimization.
Regions with $\alpha(\vx)\!\approx\! 0$ are essentially never sampled, but we must clearly visit them \emph{sometimes} to even consider the possibility of placing a surface there.

Recall the many-worlds derivative transport in \autoref{eqn:manyworlds-deriv}:
\begin{equation*}
    \partial_\pi \Li(0)
    =\int_0^s  \Big(  \underbracket{\partial_\pi \occupancy(t) \, [\LoS(t) - \Lobg(s)]}_{\text{(i)}} +
    \underbracket{\occupancy(t) \, \partial_\pi \LoS(t)}_{\text{(ii)}}  \Big) \dt.
\end{equation*}
Term (i) of this expression states that the derivative of occupancy along the ray is proportional to $\LoS(t)-\Lobg(s)$, which is an unknown quantity that must be estimated.
Given that there are \emph{no remaining known factors} that could be used to sample $t$, the best strategy for this derivative is to \emph{uniformly} pick points along the ray.

This isn't a new discovery: \citeauthor{NimierDavid2022Unbiased}~\shortcite{NimierDavid2022Unbiased} discuss the same issue in inverse volume rendering and address it by introducing generalizations of volume trackers that sample proportional to pure transmittance $T(t)$ instead of the default extinction-weighted transmittance $\mut(t) T(t)$.
Many-worlds transport has no transmittance, so the solution ends up being even simpler: a uniform sampling strategy suffices.

\end{document}